\PassOptionsToPackage{square,comma,numbers,sort&compress}{natbib}
\documentclass{article}
\pdfoutput=1

\usepackage{arxiv}

\usepackage[utf8]{inputenc} 
\usepackage[T1]{fontenc}    
\usepackage{hyperref}       
\usepackage{url}            
\usepackage{booktabs}       
\usepackage{amsfonts}       
\usepackage{nicefrac}       
\usepackage{microtype}      

\usepackage{xcolor} 
\usepackage{verbatim} 
\usepackage{pgf}
\usepackage{multirow} 
\usepackage{amsmath} 

\newcommand{\idest}{{\it i.e.}}
\newcommand{\exemp}{{\it e.g.}}
\newcommand{\etc}{{\it etc.}}
\newcommand{\etal}{{\it et al.}}

\newcommand{\nf}[1]{\texttt{#1}} 

\newcommand{\txtsub}[1]{$_\text{#1}$} 
\newcommand{\txtsob}[1]{$^{\text{#1}}$} 

\title{Evaluating the Complementarity of Taxonomic Relation Extraction Methods Across Different Languages}

\author{
  Roger Granada$^1$\thanks{This study was financed in part by the Coordena\c{c}\~{a}o de Aperfei\c{c}oamento de Pessoal de N\'ivel Superior - Brasil (CAPES) - Finance Code 001.} \And Renata Vieira$^1$ \And Cassia Trojahn$^2$ \And Nathalie Aussenac-Gilles$^3$ \AND\\
  $^1$Pontifical Catholic University of Rio Grande do Sul (PUCRS) \\
  $^2$Universit\'e Toulouse - Jean Jaur\`es \& IRIT \\
  $^3$IRIT -- CNRS \\
  \texttt{roger.granada@acad.pucrs.br}, \texttt{renata.vieira@pucrs.br} \\
  \texttt{\{cassia.trojahn, nathalie.aussenac-gilles\}@irit.fr}
}

\begin{document}
\maketitle

\begin{abstract}
Modern information systems are changing the idea of ``data processing'' to the idea of ``concept processing'', meaning that instead of processing words, such systems process semantic concepts which carry meaning and share contexts with other concepts. 
Ontology is commonly used as a structure that captures the knowledge about a certain area via providing concepts and relations between them. 
Traditionally, concept hierarchies have been built manually by knowledge engineers or domain experts. 
However, the manual construction of a concept hierarchy suffers from several limitations such as its coverage and the enormous costs of its extension and maintenance. 
Furthermore, keeping up with a hand-crafted concept hierarchy along with the evolution of domain knowledge is an overwhelming task, being necessary to build concept hierarchies automatically.
Ontology learning, usually referred to the (semi-)automatic support in ontology development, is usually divided into steps, going from concepts identification, passing through hierarchy and non-hierarchy relations detection and, seldom, axiom extraction. 
It is reasonable to say that among these steps the current frontier is in the establishment of concept hierarchies, since this is the backbone of ontologies and, therefore, a good concept hierarchy is already a valuable resource for many ontology applications. 
A concept hierarchy is represented with a tree-structured form with specialization and generalization relations between concepts, in which lower-level concepts are more specific while higher-level concepts are more general. 
The automatic construction of concept hierarchies from texts is a complex task and since the 1980 decade, much work have been proposing approaches to better extract relations between concepts. 
These different proposals have never been contrasted against each other on the same set of data and across different languages. 
Such comparison is important to see whether they are complementary or incremental, also we can see whether they present different tendencies towards recall and precision, \idest, some can be very precise but with very low recall and others can achieve better recall but low precision. 
Another aspect concerns the variation of results for different languages. 
This paper evaluates these different methods on the basis of hierarchy metrics such as density and depth, and evaluation metrics such as Recall and Precision. 
The evaluation is performed over the same corpora, which consists of English and Portuguese parallel and comparable texts. 
The output of seven methods are evaluated automatically. 
Results shed light over the comprehensive set of methods that are the state of the art according to the literature in the area.
\end{abstract}

\section{Introduction}
\label{sec:introduction}

In the past years we observe an increasing number of work addressing taxonomic relation extraction between terms. 
These work usually propose new methods or adapt old methods (\exemp, using patterns for extracting relations from the Web) and apply them in specific tasks to observe their performance. 
As these work usually use different data sets, their results are not comparable. 
For instance, the precision achieved by a method using Wikipedia can not be compared with the precision achieved by another method using the New York Times texts. 
For the best of our knowledge, these different methods have never been contrasted against each other on the same set of data and across different languages. 
As they have never been contrasted, a number of questions rises:

\begin{enumerate}
\item Is there a method that outperform all other methods?
\item If changing the language, the method performs equally?
\item All methods generate similar taxonomies?
\item Are results generated by different methods complementary or dissimilar?
\end{enumerate}

Thus, this paper addresses the answers for these questions by developing a set of methods that are commonly used in the literature of the area. 
For the first question we performed evaluations, analyzing precision, recall and f-measure scores of each method. 
For the second question we performed experiments using English and Portuguese corpora. 
For the third question we analyze the learned taxonomies on the basis of hierarchy metrics such as width and depth. 
For the last question we analyze the complementarity of the results generated by each method.

This paper is organized as follows.
\textit{Section \ref{sec:literature_review}: Literature Review on Taxonomic Relation Extraction}  describes methods for extracting hierarchical or taxonomic relations from text corpora.
\textit{Section \ref{sec:taxonomy_evaluation}: Approaches for Taxonomy Evaluation}  presents the main criteria used by related work to evaluate automatically constructed hierarchies. 
\textit{Section \ref{sec:material_methods}: Materials and Methods} describes the methodology  for developing and evaluating models that extract taxonomic relations from text corpora in two different languages, as well as the resources used and the process of evaluation. 
\textit{Section 5: Evaluation of Taxonomic Relation Extraction Methods}  presents a series of experiments aiming to evaluate the methods presented in Section 4. 
In order to verify the quality of the extracted relations and indirectly the quality of the methods that generated such relations, automatic evaluations are performed. 
Besides the analysis of the results of all methods, we also present an analysis on the characteristics of the taxonomies generated by each method on the basis of hierarchy metrics such as depth and width, and finally, an analysis on the complementarity of the methods is shown and discussed. 
The complementarity indicates whether relations generated by one method are complementary or similar to the relations generated by other methods.
\textit{Section 6: Conclusions and Future Work}  presents our conclusions and a number of directions for further work.

\section{Literature Review on Taxonomic Relation Extraction}
\label{sec:literature_review}

This section describes methods for extracting hierarchical (taxonomic) relations from text corpora, \idest, relations between a more specific and a more general term. 
Seeing through an ontology learning perspective, the methods presented in this chapter do not differentiate the specific type of relation between two concepts, indicating a class inclusion (car is-a vehicle) and between a concept instance and its superordinate concept, indicating class membership relation (Brazil is-instance-of country). 
Throughout the text we will use \nf{is-a} for both types of relations.

In this work, we consider mainly methods that extract relations based on little or no supervised algorithms, \idest, methods based on rules and methods that use the distribution of the words as an indicative of a taxonomic relationship. 
We did not include, supervised algorithms, which rely on  externally supplied instances to produce general hypotheses and then make predictions about future instances \cite{Kotsiantis2007}. 
As we do not have manually annotated data to use in the training step, this kind of algorithm is out of the scope of this paper. 

\subsection{Methods Based on Lexico-Syntactic Patterns}
\label{subsec:patterns}

The idea of learning taxonomic relations from texts by using lexico-syntactic patterns in the form of regular expressions has been introduced by Hearst \cite{Hearst1992,Hearst1998}. 
The main idea underlying using patterns is that even if one has never encountered a term, he can infer its semantic relation. 
For example, consider the following phrase ``The bow lute, such as the Bambarandang, is plucked and has an individual curved neck for each string.'' and the pattern ``\nf{NP such as \{NP ,\}* \{or$|$and\} NP}'', meaning that a noun phrase (\nf{NP}) must be followed by the words ``such'' and ``as'', which must be followed by an NP or by a list of noun phrases separated by comma, and containing before the last NP ``or'' or ``and''. 
From this example, even if one have never encountered the term ``Bambarandang'', it can be inferred that ``Bambarandang'' is a kind of ``bow lute''. 
Thus, the underlying idea of using lexico-syntactic patterns is very simple: to define regular expressions that capture expressions and to map the results of the matching expression to a taxonomic structure between terms. 

Patterns proposed by Hearst~\cite{Hearst1992,Hearst1998} were initially developed for English, but they have been widely spread to other languages such as Japanese~\cite{AndoEtAl2004}, Dutch~\cite{SangHofmann2007}, Turkish~\cite{YildizYildirim2012}, French~\cite{MorinJacquemin1999} and Portuguese~\cite{Basegio2007}. 
Taxonomic relations can be extracted fairly accurately using syntactic patterns \cite{NastaseEtAl2013}. 
In contrast, these patterns are usually brittle and may not occur very often in a corpus. 
Although approaches relying on lexico-syntactic patterns have a reasonable precision, their recall is very low \cite{BuitelaarEtAl2005,Cimiano2006}. 

An approach to minimize the drawbacks of low coverage has been proposed by Pantel and Pennacchiotti~\cite{PantelPennacchiotti2006} that apply the patterns proposed by Hearst \cite{Hearst1992,Hearst1998} using the Web as a big corpus. 
In the same direction, Ponzetto and Strube \cite{PonzettoStrube2011} attached the patterns presented by Hearst \cite{Hearst1992,Hearst1998} in a method for building a taxonomy based on the content of Wikipedia structure. 
In their approach, the semantic relations between categories are labeled either \nf{is-a} or \nf{not-is-a} using methods based on connectivity in the network and lexico-syntactic matching. 
In order to improve the precision of terms extracted after applying the patterns proposed by Hearst~\cite{Hearst1992,Hearst1998}, Cederberg and Widdows \cite{CederbergWiddows2003} use Latent Semantic Analysis (LSA) to filter out terms that are not semantically related. 
When applying LSA, they intend to reduce the rate of error of the initial pattern-based hyponymy extraction. 
Even with drawbacks, much work keep using the patterns proposed by Hearst to extract hyponym relations between terms, sometimes combining them with other techniques such as head-modifier \cite{LefeverEtAl2014,Lefever2015}, clustering \cite{Caraballo1999,DegeratuHatzivassiloglou2004} or LSA \cite{CederbergWiddows2003}, sometimes using them on the Web to extract contexts \cite{Pasca2004} or class instances \cite{EtzioniEtAl2005,KozarevaEtAl2008}.

\subsection{Methods Based on Head-Modifier Detection}
\label{subsec:headmodifier}

According to Radford \cite{Radford1997}, the head of a phrase is the grammatically most important word in the phrase, since it determines the nature of the overall phrase. 
The changing in the sense of the head can be classified according to the particle attached to the head. 
In a term composed by affixation, an affix is added to a word forming a new word, \exemp, in ``preprocessor'' the word ``processor'' is the head and ``pre'' is an affix. 
In a compounding term, two or more words come together to generate a new meaning. 
The compounding term can be composed by a single word with the merging of two or more words, or be composed by two or more elements. 
When compounding terms are consisted by two or more elements, it is claimed that the arrangement of these elements reflects the kind of information being conveyed. 
The main element, as known the head, identifies the semantic category to which the whole term belongs, and the other elements distinguish these members from other members of the same category. 
Thus, the whole compound term is related to the head as a hyponym. 

\begin{figure}[b!]
    \centering
    \includegraphics[width=0.8\textwidth]{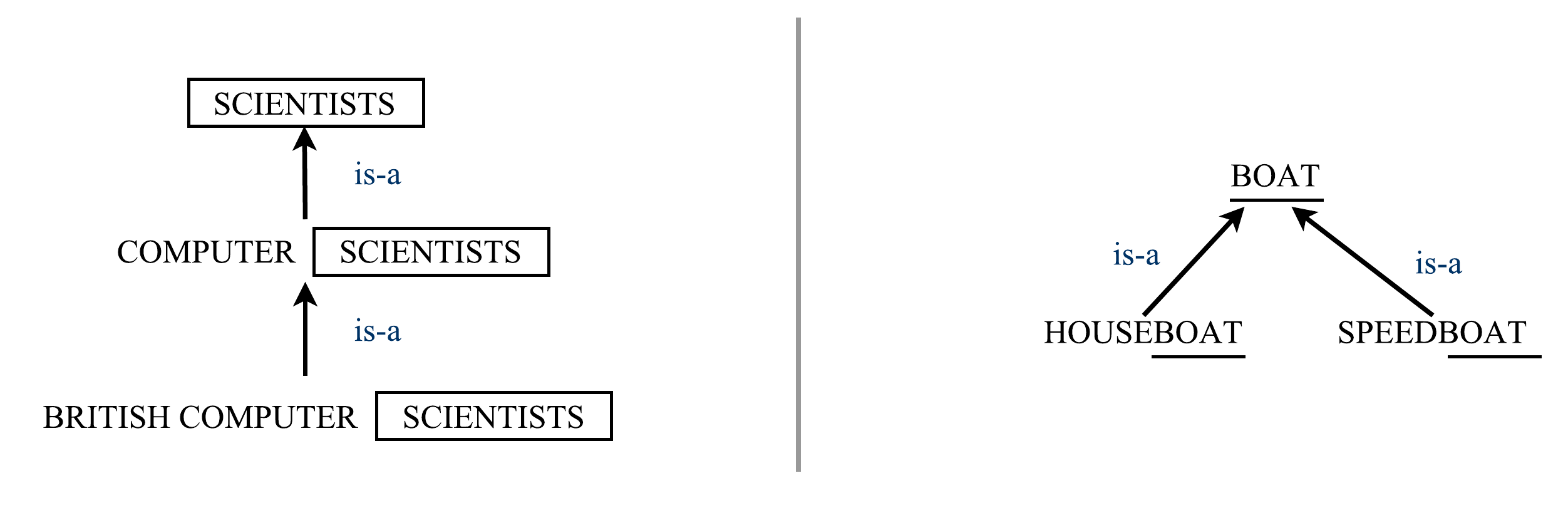}
    \caption{Example of compounding terms containing one single word and two or more elements.}
    \label{fig:compounding}
\end{figure}

Following the example in Figure \ref{fig:compounding}, the term ``Computer~Scientists'' has as head ``Scientists'' which can be transformed into its hypernym, \idest, ``Computer~Scientists'' is a kind of ``Scientist'', and ``British~Computer~Scientists'' has the head ``Computer~Scientists'' which can be its hypernym, \idest, ``British~Computer~Scientists'' is a kind of ``Computer~Scientists''. 
In contrast, compounding terms containing a single word have the head as part of the word. 
Thus, in constructions such as ``Houseboat'' and ``Speedboat'' the head element is ``Boat'', and may therefore be viewed as the hypernym. 
The affixed terms ``Houseboat'' and ``Speedboat'' are hyponyms of ``Boat'', \idest, a kind of ``Boat''. 
The modifiers ``house'' and ``speed'' act to distinguish the members of the set of hyponyms \cite{HippisleyEtAl2005}.

\begin{figure}[t!]
    \centering
    \includegraphics[width=0.70\textwidth]{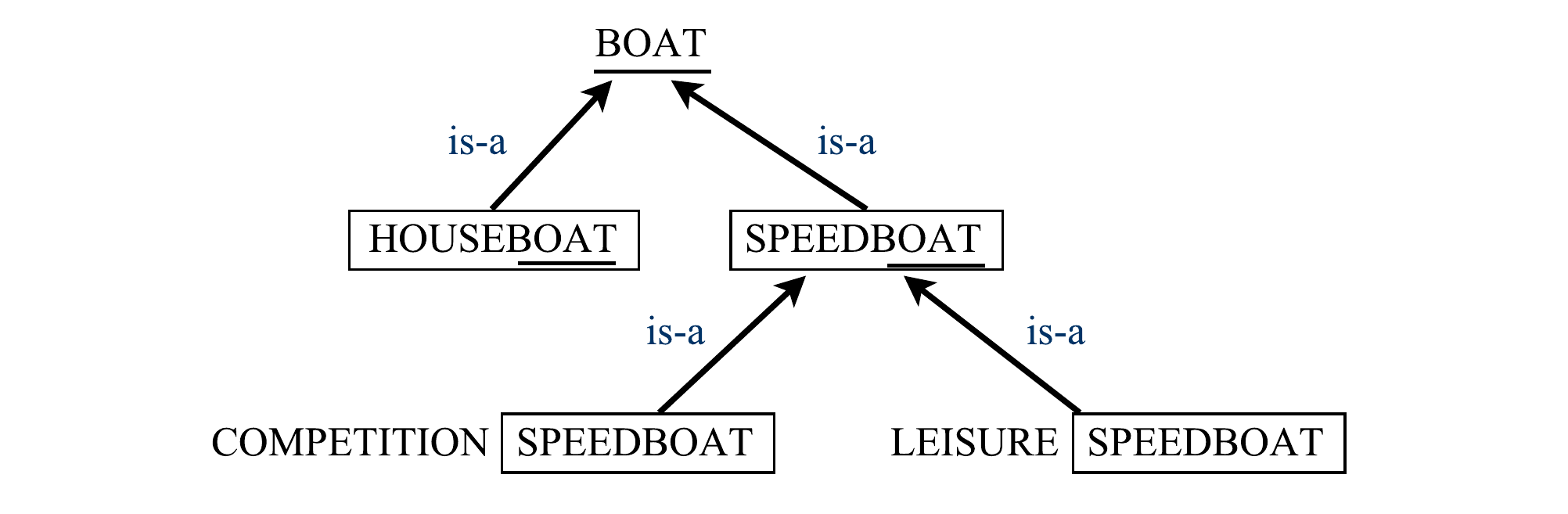}
    \caption{Example of head-modifier relations containing mixed compounding terms.}
    \label{fig:boat_hyponyms}
\end{figure}

Following this idea, it is also possible to have a mix of compounding terms. 
Figure \ref{fig:boat_hyponyms} presents the hierarchy using head-modifier to the term ``Boat'', where the terms ``Competition~Speedboat'' and ``Leisure~Speedboat'' are a kind of ``Speedboat'', as well as ``Speedboat'' is a kind of ``Boat''. 
Finally, the terms ``Competition~Speedboat'' and ``Leisure~Speedboat'' are connected to ``Boat'', through the term ``Speedboat''.

Terms containing three or more words (\exemp, ``film~society~committee'') could be linked to its hypernym (\exemp, ``society~committee''), which is also linked to its own hypernym (\exemp, ``committee''), and thus creating natural sub-levels. 
However, the chunking of such terms can be ambiguous. 
Using the example above, the term ``film~society~committee'' can be decomposed into two ways: [film [society~committee]] or [[film~society] committee]. 
Note that the bracketing is important and it indicates the subconstituency of the term, and therefore its derivational history \cite{HippisleyEtAl2005}. 
The former example can be interpreted as ``There exist committees, some of which are society committees. There are range of these, including society committees whose interest is film.'' and the latter be interpreted as ``There exist committees, some of which are film society committees.''. 

In order to eliminate the ambiguity many techniques have been applied, such as using the term frequencies. 
For example, the frequency of [society~committee] can be compared with that of [film~society] to give the likelihood of the candidate bracketing. 
Velardi \etal~\cite{VelardiEtAl2001} use Mutual Information \cite{Fano1961} and Dice factor \cite{SmadjaEtAl1996} to the terminology extraction and then calculate the Domain Relevance (DR) for term disambiguation. 
Vossen~\cite{Vossen2001} extracts the most likely chunking by looking for the most salient head and decomposing the remaining words into modifiers. 

As pointed out by Buitelaar \etal~\cite{BuitelaarEtAl2004}, in many languages the morphological system is very rich and enables the construction of semantically complex compound words, \exemp, the German word ``Kreuzbandverletzung'' corresponds to three English words: ``cruciate'', ``ligament'' and ``injury'' that are combined into a single word containing the meaning of all words. 
In order to deal with these complex compounds, Buitelaar \etal~\cite{BuitelaarEtAl2004} and Sintek \etal~\cite{SintekEtAl2004} present an environment for the integration of linguistic analysis in ontology engineering through the definition of mapping rules. 
One of this mapping rules splits nouns into its chain of elements and uses the head of the chain to build the hypernym/hyponym relations between terms. 

Lopes \cite{Lopes2012} presents a process to the automatic extraction of concept hierarchies from domain corpora in Portuguese. 
This process first extracts domain terms represented as noun phrases containing simple or compound terms. 
From the selected terms, the head of each noun phrase assumes the role of hypernym and the whole noun phrase the role of hyponym. 
Noun phrases that contain inner terms are also expanded in a taxonomic structure. 

Other work combine head-modifier with other approaches in order to improve the relation extraction. 
For example, Espinosa \etal~\cite{EspinosaEtAl2015} combine a head-modifier approach with a supervised method in order to extract hypernyms when the classifier has low confidence to tag tokens as possible hypernyms. 
They call this process \emph{hypernym decomposition} and intends to generate deeper paths from a term and its hypernym by successively decomposing the hypernyms. 
For instance, consider the candidate relation (``term''$\rightarrow$\emph{hypernym}) in ``whisky''$\rightarrow$``distilled alcoholic beverage'', the hypernym is decomposed into ``distilled alcoholic beverage''$\rightarrow$``alcoholic beverage''$\rightarrow$``beverage'', generating longer hypernymy paths.

\subsection{Methods Based on Distributional Analysis}
\label{subsec:distributional_analysis}

Many methods that process data involve semantic models, also known as corpus-based semantic models, semantic spaces, word spaces, or distributional similarity models (DSMs). 
Such models usually rely on some version of the distributional hypothesis \cite{Harris1954} which states that words that occur in the same contexts tend to have similar meanings. 
In other words, the degree of semantic similarity between two words is related to the degree of overlapping among their contexts. 
A better understanding about semantic similarity is presented by Lemaire and Denhi\'ere~\cite{LemaireDenhiere2006} who point out that it could be viewed as an association of two terms, that is, the mental activation of one term when another term is presented. 

Usually DSMs vary according to the aspects of meaning they are designed to model. 
According to Medin \etal~\cite{MedinEtAl1990} there are two types of similarity, the attributional similarity, \idest, a correspondence between attributes, and the relational similarity, \idest, a correspondence between relations. 
As noted by Grefenstette~\cite{Grefenstette1994}, attributional similarity is typically addressed by word collocates, that is, words that co-occur more often than would be expected by chance. 
Thus, words that share many collocates denote concepts that share many attributes. 
For example, the contextual similarity between ``automobile'' and ``car'' is very high because they co-occur with ``accelerate'', ``break'', ``color'' and other words. 
On the other hand, two word pairs are analogous when they have a high degree of relational similarity (\exemp, ``traffic'' is to ``street'' as ``water'' is to ``riverbed''). 

Sanderson and Croft \cite{SandersonCroft1999} present a measure based on the probabilities of terms co-occurrence. 
The main idea is that a term $x$ is the hypernym of $y$ if the conditional probability of the term $x$ is greater than the conditional probability of the term $y$ and above a threshold.
In their work the threshold was set to 1, but after some experiments the authors noticed that many terms were not included because a few occurrences of the hyponym $y$, did not co-occur with $x$. 
Thus, after some analysis of hypernym-hyponym pairs, the threshold was set to 0.8. 
Later on, Njike-Fotzo and Gallinari \cite{FotzoGallinari2004} extended this idea replacing the conditional probability as basic co-occurrence evidence by the Expectation-Maximization (EM) algorithm \cite{GuptaChen2011}. 
Such approach is based on log-likelihood indices and it is less accurate, however, it is faster to compute, thus, being applicable to larger text sources. 

\begin{figure}[htb!]
    \centering
    \includegraphics[width=0.75\textwidth]{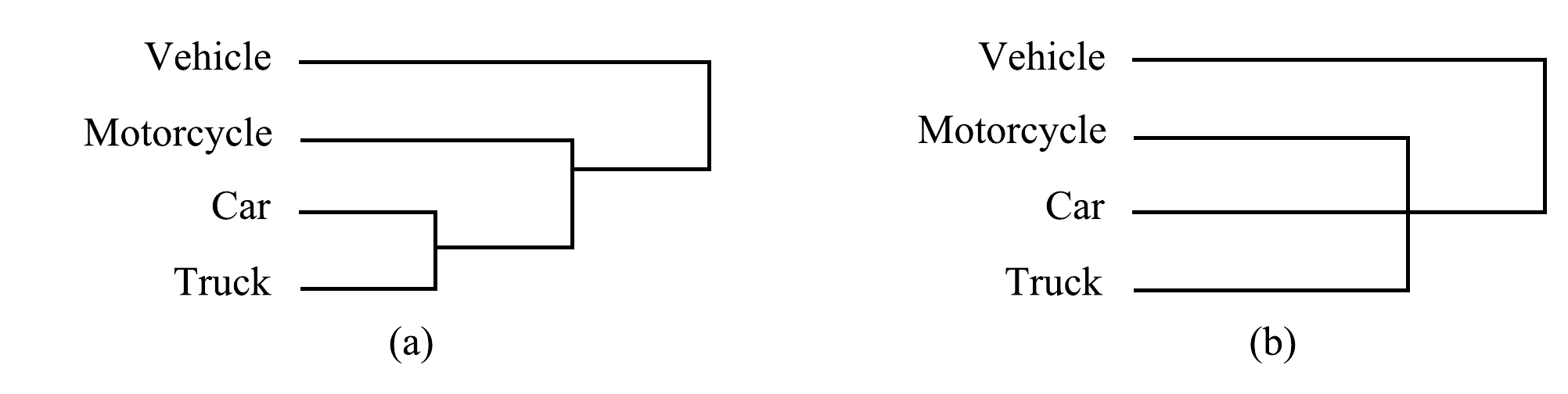}
    \caption{Representation of coordinate terms by a binary tree.}
    \label{fig:binarytree}
\end{figure}

Associating similar words, Caraballo~\cite{Caraballo1999} creates hypernym-labeled noun using hierarchical clustering. 
Initially, all nouns are hierarchically clustered by an agglomerative bottom-up clustering of vectors having conjunction and appositives as features is generated. 
Since traditional hierarchical clustering algorithms construct binary trees, Caraballo applies a step to compress the generated tree by looking when an internal node is unlabeled, meaning that a hypernym could not be found to describe its descendant nouns. 
Since there is a labeled node with a hypernym above the current node, delete the current node and connect the children of this node to the hypernym. 
This step is important because binary branches may not correctly describe the data structure. 
For example, Figure \ref{fig:binarytree} presents the structure generated by a traditional hierarchical clustering algorithm ($a$), and how the tree should be structured ($b$), since the words ``Brazil'', ``France'' and ``USA'' are coordinate terms, \idest, terms that share the same hypernym (\exemp, ``country''). 

Liu \etal~\cite{LiuEtAl2012} apply a multi-branched tree algorithm during the hierarchical clustering process, avoiding the problem of generating a binary tree. 
In order to build the multi-branched tree, they adopted a deterministic, agglomerative approach proposed by Blundell \etal~\cite{BlundellEtAl2010} called Bayesian Rose Tree (BRT). 
This approach explores a Bayesian hierarchical clustering algorithm that can produce trees with arbitrary branching structure at each node. 

Similarly to Caraballo \cite{Caraballo1999}, De Knijff \etal~\cite{DeKnijffEtAl2013} build a taxonomy using hierarchical clustering methods. 
In that work De Knijff \etal~claim that this type of method may generate hierarchies with terms containing multiple potential parents. 
Hence, they choose one potential parent to maintain the hierarchical tree structure. 
The decision is based on a score calculated for each potential parent, taking into account the distance between the target term and the list of ancestors parents. 

According to Buitelaar \etal~\cite{BuitelaarEtAl2004} methods that rely on raw data or frequency counting may lead to data sparseness. 
In order to overcome this problem, approaches based on dimension-reduction techniques like Latent Semantic Analysis (LSA) \cite{LandauerDumais1997} should be applied on term by context matrices. 
Gamallo and Bordag \cite{GamalloBordag2011} explain that LSA can identify relations between words that do not co-occur directly in the corpus, \idest, words that do not share the same contexts.  
These co-occurrences can be obtained by means of applying Singular Value Decomposition (SVD) methods. 
Thus, SVD methods try to represent a more abstract and generic word space which tries to capture higher-order associations by inducing a latent (hidden) structure that does not rely on word co-occurrences attested in the corpus.

Word embeddings is another distributed word representation that represent words with dense, low-dimensional and real valued vectors \cite{MikolovEtAl2013wordrepresentation}. 
In word embedding models, also known as distributed word representations, the words are mapped to vectors ($v$) of real numbers in a low-dimensional space relative to the vocabulary size (\exemp, using dimensionality reduction on the word co-occurrence matrix). 
Word embedding models have shown to preserve the linguistic regularities which can be recovered by simple vector offsets between the vector embeddings of each word \cite{MikolovEtAl2013wordrepresentation}. 
For example, $v(king) - v(queen) \approx v(man) - v(woman)$, where $v(w)$ is the embedding of the word $w$. 

Recently, Tan \etal~\cite{TanEtAl2015} use vector space word embedding models for hypernym-hyponym extraction. 
Along with all vectors for nouns, Tan \etal~create a vector that represents the single-tokenized ``\nf{is-a}'' phrase. 
All vectors are created using skip-gram model phrasal \emph{word2vec} neural net \cite{MikolovEtAl2013wordrepresentation}. 
Finally, hypernyms are obtained by the multiplication of the hyponym vector by the ``\nf{is-a}'' vector. 
For example, in order to obtain the hypernym of ``goldfish'', the vector $v$(goldfish) is multiplied by the vector of ``\nf{is-a}'' $v$(is-a) which should be similar to the cross product $v$(fish), $v$(goldfish)$\times v$(is-a) $\approx v$(fish). 

Pocostales~\cite{Pocostales2016} identifies hypernyms by adding a vector offset to the corresponding hyponym word embedding. 
This vector offset is obtained as the average offset between 200 pairs of hyponym-hypernym in the same vector space. 
In order to create the vector space representations of words, Pocostales use the GloVe~\cite{PenningtonEtAl2014} log-bilinear model trained on the Wikipedia corpus. 
Since the vector generated by join the offset vector with the hyponym vector will rarely match the exact vector of the hypernym, the Cosine similarity is applied to identify the most similar vector, and thus, the most probable hypernym. 
When performing experiments, Pocostales concludes that the diversity involved in the complex hypernym-hyponym relations cannot easily be captured by a simple vector offset mean. 

\subsection{Methods based on Distributional Inclusion}
\label{subsec:distributional_inclusion}

Hyponymy relation is transitive, \idest, a term $u$ is related to a term $v$, and $v$ is in turn related to a term $w$, then $u$ is also related to $w$; and asymmetrical, \idest, $u$ is a kind of $v$, but $v$ is not a kind of $u$. 
For example, ``potato'' is a kind of ``vegetable'' and ``vegetable'' is a kind of ``plant''. 
Thus, by transitivity ``potato'' is a kind of ``plant'' and by asymmetry ``plant'' is not a kind of ``potato''.

These relations generate a hierarchical semantic structure where the hyponym is below its superordinate, and where the hyponym inherits all the features of the more generic concept and adds at least one feature that distinguishes it from its superordinate and from any other hyponyms of that superordinate \cite{MillerEtAl1990}. 
For example, ``mapple'' is an hyponym of ``tree'' because it inherits the properties of the latter but is distinguished from the other trees by the hardness of its wood, shape of its leaves \etc. 
On the other hand, Weeds \etal~\cite{WeedsEtAl2004} verified that distributional generality is correlated with semantic generality, \idest, hypernyms tend to occur in a larger variety of contexts than hyponyms. 
In this sense, the asymmetry is captured by means of co-occurrence retrieval \cite{WeedsWeir2003}, where a word has higher recall and lower precision when compared with its hyponyms'~co-occurrences, and higher precision and lower recall when compared with its hypernyms'~co-occurrences. 

Observing the distributional generality of words in the work of Weeds \etal, Geffet and Dagan~\cite{GeffetDagan2005} point out that sources of noise in the Weeds'~work may be influenced in the results, not showing a great improvement when using feature vectors. 
The authors claim that the quality of similarity scores is often biased by inaccurate feature weights. 
Thus, they propose a recalculation on weighted vectors taking into account the set of most similar words generated by the Lin's measure~\cite{Lin1998clustering}. 

Using the new weighted vectors, Geffet and Dagan propose the Distributional Inclusion Hypothesis. 
The hypothesis says that if the meaning of a word $u$ entails another word $v$, then it is expected that all the typical contexts (features) of $u$ will occur also with $v$. 
In other words, if a term $u$ is semantically narrower than term $v$, then a significant number of salient distributional syntactic features of $u$ is also included in the feature vector of $v$. 

Szpektor and Dagan \cite{SzpektorDagan2008} proposed a balanced version of the Weeds \etal~\cite{WeedsEtAl2004} work, combining the precision of achieved by Weeds~\etal with the Lin's measure by taking their geometric average. Thus, Lin's measure \cite{Lin1998clustering} works on penalizing vectors containing few features. 
Clarke \cite{Clarke2009} formalized the idea of distributional generality and computes the entailment between two words using a variation of Weeds \etal~\cite{WeedsEtAl2004} measures. 
It differs from the one proposed by Weeds \etal~because it reduces the weight of included features if they have lower weight within the vector of the broader term. 

Lenci and Benotto \cite{LenciBenotto2012} expand the idea of Geffet and Dagan \cite{GeffetDagan2005} and explore the possibility of identifying hypernyms in Distributional Similarity Models (DSMs) using directional (or asymmetric) similarity measures. 
They propose a new measure that takes into account not only the inclusion of the features of $u$ in $v$, but also the non-inclusion of the features $v$ in $u$. 
The idea is that, if $v$ is a semantically broader term of $u$, then the features of $u$ are included in the features of $v$, but crucially the features of $v$ are not included in the features of $u$.

Kotlerman \etal~\cite{KotlermanEtAl2010} craft a measure that is optimized to capture a relation of feature inclusion between terms while using the relative relevance of features. 
This measure applies the IR evaluation method of \emph{Average Precision} in order to identify the feature inclusion, while uses the symmetric similarity measure of Lin~\cite{Lin1998clustering} to penalize low frequency words. 
The idea is that the score increases with a larger number of features shared by $u$ and $v$, while giving higher weight to highly ranked features of the narrower term.

Santus \etal~\cite{SantusEtAl2014} use an entropy-based measure named \emph{SLQS} for the unsupervised identification of taxonomic relations in DSMs. 
\emph{SLQS} is grounded on the idea that contexts of hypernyms contain less information than the most contexts of its hyponyms. 
For example, contexts like ``has fur'' and ``bark'' are likely to co-occur with a smaller number of terms than ``move'' and ``eat''. 
Hence, \emph{SLQS} uses entropy \cite{Shannon1948} as an estimate of context informativeness.

\section{Approaches for Taxonomy Evaluation}
\label{sec:taxonomy_evaluation}

Evaluating hierarchical structures is still a hard task, being difficult even for humans, since there is no unique way to generate a correct structure and sometimes different taxonomies may model a domain equally well. 
Evaluations can be manually or automatically performed. 
The former is based on manual evaluation where domain experts assess the structure and its relations. 
The latter compares the automatically extracted taxonomy against a gold standard by comparing both structures and relations, or checking the adequacy within an application when using such structure. 

Manual evaluation is usually performed on novel or specific domains where a gold standard is not available. 
Furthermore, a manual evaluation allows to assess not only the taxonomic relations between terms, but also the quality of the whole structure in a detailed view. 
On the other hand, the manual evaluation is rarely performed on the whole taxonomy, but in a subset of all relations. 
For domain experts deciding whether or not a term belongs to the domain is more or less feasible. 
Furthermore, deciding the quality of a taxonomic relation is a more complex task. 
As mentioned by Velardi \etal~\cite{VelardiEtAl2013}, when annotators where asked to blindly produce a taxonomy from a given set of terms, they struggled with the domain terminology and produced a quite messy organization. 
Manual evaluation also has the drawbacks of being a time consuming task and depending on the availability of domain experts (for some domains it may be quite hard to find experts that can dedicate time for evaluating). 
Due to these factors, it is highly desired to have a gold standard to support an automatic validation. 

While the manual evaluation is a laborious task, the automatic evaluation gives us the possibility of testing a set of relations, comparing results against a gold standard, \idest, a hierarchy manually constructed by domain experts that serves as reference of domain terms and taxonomy (\exemp~WordNet~\cite{Fellbaum1998}). 
In such evaluation the quality of the identified hierarchy is expressed by its similarity to the gold standard hierarchy, measuring the ability to reproduce the relations between pairs of words. 
The problem of using such structure is that imperfections in the gold standard will impact in the results. 
This kind of imperfections vary from missing terms, \idest~the gold standard has low coverage and it does not contain all terms contained in the generated hierarchy, to missing relations between terms. 
For example, WordNet has an entry for ``Robert De Niro'' (United States film actor) as an instance of ``actor'', a subclass of ``performing artist'', ``entertainer'', ``person'' and ``being'', but not a subclass of ``man''. 
Another problem of the automatic evaluation refers to the availability of gold standards for certain domains. 
Usually, the results of automatic evaluation are in terms of precision, recall and F-measure, and denote how well a method ``mimic'' the gold standard. 

Table \ref{tab:automatic_eval} summarizes the work presented in this paper that automatically evaluate the results generated by hierarchical relations extractions from texts, where \nf{P} denotes precision, \nf{R} recall and  \nf{F} F-measure, \nf{F\txtsub{t}}, \nf{P\txtsub{t}} and \nf{R\txtsub{t}} are relative to the tourism gold standard, \nf{F\txtsub{f}}, \nf{P\txtsub{f}} and \nf{R\txtsub{f}} are relative to the finance gold standard, \nf{F\txtsub{B}} is the best F-measure, \nf{MAP\txtsub{+NE}} means the Mean average precision with additional Named Entities, \nf{MAP\txtsub{+Feat}} is the Mean average precision with additional features, \nf{AP} denotes average precision, \nf{Cv} represents the coverage, \nf{Nv} the novelty, and \nf{EC} the extra coverage, \nf{ACC\txtsub{1}} the accuracy of the Test-I and \nf{ACC\txtsub{2}} that of the Test-II. 
Although the evaluation measures the similarity with a gold standard, the results are usually influenced by the imperfections presented in those gold standards. 
Lexical databases such as WordNet \cite{Fellbaum1998} or hand-crafted lists containing terms and their semantic relation are commonly used as gold standards. 

\begin{table*}[htb]
 \scriptsize
 \caption{Work that uses automatic evaluation for the extracted relations.}
 \label{tab:automatic_eval}
 \centering
 \resizebox{\textwidth}{!}{\begin{tabular}{clll}
  \hline\noalign{\smallskip}
 {\bf Reference}            & {\bf Resources used}            & {\bf Evaluation}                                  & {\bf Results}              \\
  \noalign{\smallskip}\hline\noalign{\smallskip}
\cite{Hearst1992}           & Grolier's American Academic     & 106 relations of the ``NP such as LNP'' pattern   & P=57.5\%                   \\
                            & Encyclopedia \cite{Grolier1998} & using WordNet v.1.1 \cite{MillerEtAl1990}         &                            \\
                                                                                                                  &&&                          \\
\cite{WeedsEtAl2004}        & British National Corpus (BNC)   & 20,415 pairs of BNC using WordNet v.1.6           & P=71\%                     \\
                                                                                                                  &&&                          \\
\cite{GeffetDagan2005}      & Reuters RCV1 corpus             & 400 annotated pairs from Reuters corpus           & P=70\%                     \\
                                                                                                                  &&&                          \\
\cite{CimianoEtAl2005}      & LonelyPlanet                    & Semantic Cotopy and Taxonomy overlap              & F\txtsub{t}=40.52\%  \\
                            & All in all                      & using Tourism ontology \cite{MaedcheStaab2002}    & P\txtsub{t}=29.33\%  \\
                            & BNC corpus                      &                                                   & R\txtsub{t}=65.49\%  \\
                            & Reuters 1987\footnote{http://www.daviddlewis.com/resources/testcollections/reuters21578/} & Finance ontology \cite{StaabEtAl1999} & F\txtsub{f}=33.11\%  \\
                            &                                 &                                                   & P\txtsub{f}=29.93\%  \\
                            &                                 &                                                   & R\txtsub{f}=37.05\%  \\
                                                                                                                  &&&                          \\
\cite{KotlermanEtAl2010}    & Reuters RCV1 corpus             & 3,772 annotated pairs \cite{Zhitomirsky-GeffetDagan2009}  & AP=47\%                    \\
                            &                                 &                                                           & P=32\%                     \\
                            &                                 &                                                           & R=92\%                     \\
                                                                                                                  &&&                          \\
\cite{PonzettoStrube2011}   & Wikipedia (March 2008)          & ResearchCyc\footnote{http://research.cyc.com/}    & Cv=1.6\%, Nv=99.2\%, EC=28.2\%            \\
                            &                                 & WordNet v.3.0 \cite{Fellbaum1998}                 & Cv=8.7\%, Nv=99.3\%, EC=211.6\%            \\
                            &                                 & 3,500 annotated category pairs                    & P=90.3\%, R=78.5\%, F\txtsub{B}=84\%       \\
                                                                                                                  &&&                          \\
\cite{LenciBenotto2012}     & TypeDM \cite{BaroniLenci2010}   & 14,547 tuples of BLESS \cite{BaroniLenci2011}     & AP=40\%                    \\
                                                                                                                  &&&                          \\
                            &                                 &                                                   & ACC\txtsub{2}=86.98\%      \\
                                                                                                                  &&&                          \\
\cite{RiosAlvaradoEtAl2013} & LonelyPlanet                    & A gold standards associated to each corpus        & P=53\%, R=89\%, F=67\%     \\
                            & SmartWeb Football               &                                                   & P=77\%, R=83\%, F=80\%     \\
                            & Biology news                    &                                                   & P=49\%, R=56\%, F=52\%     \\
                            & Java \cite{LeeEtAl2005}         &                                                   & P=76\%, R=72\%, F=74\%     \\
                                                                                                                  &&&                          \\
\cite{SantusEtAl2014}       & ukWaC                           & BLESS \cite{BaroniLenci2011} data set             & P=87\%                     \\
                            & WaCkypedia                      &                                                   &                            \\  
                                                                                                                  &&&                          \\
\cite{LefeverEtAl2014}      & Financial Times                 & Manually created gold standard                    & P=76\%, R=67\%,            \\
                            & Reports form dredging company   & Manually created gold standard                    & P=69\%, R=58\%,            \\
                                                                                                                  &&&                          \\
\cite{Lefever2015}          & Web corpus                      & SEMEVAL 2015 - Task 17 \cite{BordeaEtAl2015}      & AP=36\%, AR=63\%, AF=39\%  \\
                                                                                                                  &&&                          \\
\cite{EspinosaEtAl2015}     & WCL Dataset \cite{NavigliVelardi2010} & SEMEVAL 2015 - Task 17 \cite{BordeaEtAl2015}& AP=7\%, AR=12\%, AF=8\%    \\
                                                                                                                  &&&                          \\
\cite{TanEtAl2015}          & Wikipedia (2015 dump)           & SEMEVAL 2015 - Task 17 \cite{BordeaEtAl2015}      & AP=20\%, AR=31\%, AF=24\%  \\
                                                                                                                  &&&                          \\
\cite{Pocostales2016}       & Wikipedia (2015 dump)           & SEMEVAL 2016 - Task 13 \cite{BordeaEtAl2016}      & AP=14\%, AR=30\%, AF=19\%  \\
  \noalign{\smallskip}\hline
  \end{tabular}}
\end{table*}

As pointed out by Velardi \etal~\cite{VelardiEtAl2013} it is not clear how to evaluate the concepts and relations not found in the gold standard. 
As these terms can be either wrong or correct, the evaluation is in any case incomplete. 
Automatic evaluation also has the drawback of choosing an adequate evaluation metric. 
For example, the taxonomy overlap computes the ratio between the intersection and union of two sets. 
Therefore this metric do not provide a structural comparison between the taxonomies, and thus, errors in the hierarchy structure are not indicated by this metric. 

Another common practice to automatically evaluate the quality of the system is to compare it against the state of the art systems, verifying their performance in some task. 
For instance, Rios-Alvarado \etal~\cite{RiosAlvaradoEtAl2013} reproduced the approaches presented by Cimiano \etal~\cite{CimianoEtAl2005} and by Jiang and Tang~\cite{JiangTang2010} in order to compare with their proposal. 
The problem in emulating the approaches proposed by someone else is that not always the resources to reproduce the system are available (\exemp, the corpus to train a machine learning system). 
Also, usually results are compared between similar approaches (\exemp, Santus \etal~\cite{SantusEtAl2014} compare their results with other work that also use directional similarity measures). 

Regarding the results obtained in each work, usually they are not comparable since the resources used in each approach are different. 
For instance, Hearst~\cite{Hearst1992} achieved 57.5\% of precision using encyclopedia texts. 
When Cederberg and Widdows~\cite{CederbergWiddows2003} applied the same method on the British National Corpus they achieved a precision of 40\%. 
Thus, it would be impossible to compare these values of precision and recall with the work presented by Cimiano \etal~\cite{CimianoEtAl2005} that obtain them using documents from finance and tourism domain. 
This difference shows how important it is to perform an evaluation of these methods using the same corpora, as we do.

\section{Materials and methods}
\label{sec:material_methods}

This section describes the methodology used in this work for the development and evaluation of the developed methods to extract taxonomic relations from text corpora for two distinct languages:  Portuguese and English. This section is separated in three parts, where Section \ref{subsec:resources} presents a brief description of each corpus and gold standards, Section \ref{subsec:preprocessing} details the pre-processing of data, Section \ref{subsec:models} describes all methods implemented for the evaluation, and Section \ref{sec:evaluations} discusses the evaluation methodology and criteria. 

\subsection{Resources}
\label{subsec:resources}

Resources are separated into corpora and gold standard details. As we are working with two different languages, resources are described as \nf{EN} for English and \nf{PT} for Portuguese.  

\subsubsection{Corpora}
\label{subsec:corpora}

According to our purposes of contrasting methods regarding the same corpora but over different languages and genres, we use, for our experiments, two parallel corpora of different nature: a collection of texts of the European Parliament, and transcribed TED talks which encompasses a vast area of knowledge. 
Here, parallel corpora can be understood as a corpus containing  direct translation of sentences, embedding the same meaning at the sentence level. 
An excerpt of aligned sentences from the Europarl parallel corpus is presented below: 

\begin{figure}[h!]
    \centering
    \includegraphics[width=\textwidth]{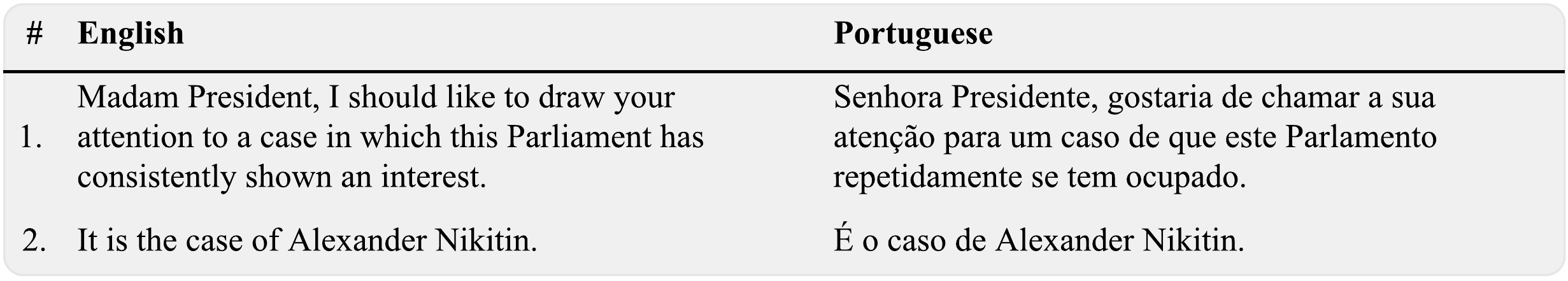}
\end{figure}

The nature of the documents may influence the number of extracted taxonomic relations, mainly using pattern methods, since such methods tend to capture relations in the definition of terms. 
The corpora used in this study are briefly described in the sequence -- note that \nf{EN}  and \nf{PT} means English and Portuguese respectively. 

\textbf{\nf{Europarl (EN-PT)}}: Europarl parallel corpus~\cite{Koehn2005} is a collection of the proceedings of the European Parliament, comprising of about 30 million words for each of the 11 official languages of the European Union: Danish, Dutch, English, Finnish, French, German, Greek, Italian, Portuguese, Spanish and Swedish. 
The bilingual corpus containing sentence aligned texts is freely available at the Statistical Machine Translation site\footnote{http://www.statmt.org/}. 
In this work we use the English-Portuguese sentence-aligned version of the parallel corpus, \idest, each sentence of the English corpus has its translation into Portuguese. 

\textbf{\nf{TED Talks (EN-PT)}}: TED is a nonprofit organization and its website\footnote{http://www.ted.com} makes available the video recordings together with subtitles provided in many languages of the best TED talks. 
Almost all talks have been translated by volunteers into about 70 other languages. 
The collection containing sentence aligned documents is provided by the Web inventory named WIT$^3$, an acronym for Web Inventory of Transcribed and Translated Talks~\cite{CettoloEtAl2012}. The collection contains 1,112 transcribed and translated talks containing topics that span the whole of human knowledge. 

Table \ref{tab:corpora_stats} informs some statistics of the corpora, where ``\nf{|D|}'' represents the number of documents in which the corpus is divided, ``\nf{|S|}'' is the number of sentences of the corpus, ``\nf{|W|}'' is the number of content words and ``\nf{|V|}'' is the size of the vocabulary of ``\nf{|W|}''. 
Size of the vocabulary is the number of words without counting repetition. 
The Europarl collections have just one document because it does not contain document borders, but a sentence aligned document. 
Thus, the English version of the corpus contains a single document with all sentences of the proceedings, while the Portuguese version contains a single document with all aligned sentences of the English version translated into Portuguese. 

\begin{table*}[ht]
    \caption{Statistics of the corpora in number of documents |D|, number of sentences |S|, number of words |W| and size of the vocabulary |V|.}
    \label{tab:corpora_stats}
    \centering
    \begin{tabular}{c|ll|r|r|r|r}
        \noalign{\smallskip}\hline\noalign{\smallskip}
        {\bf Language}      & \multicolumn{2}{l|}{{\bf Corpus}} & \multicolumn{1}{c|}{{\bf |D|}} & \multicolumn{1}{c|}{{\bf |S|}} & \multicolumn{1}{c|}{{\bf |W|}}  & \multicolumn{1}{c}{{\bf |V|}}   \\
        \noalign{\smallskip}\hline\noalign{\smallskip}
            \multirow{2}{*}{PT} & \multicolumn{2}{l|}{{Europarl}}          & 1          & 1,960,407 & 20,792,400 & 689,593  \\
                                & \multicolumn{2}{l|}{{TED Talks}}         &  1,112     & 224,049   & 1,379,163  & 43,161   \\
        \noalign{\smallskip}\hline\noalign{\smallskip}
            \multirow{2}{*}{EN} & \multicolumn{2}{l|}{{Europarl}}          & 1          & 1,960,407 & 22,159,518 & 86,367   \\ 
                                & \multicolumn{2}{l|}{{TED talks}}         & 1,112      & 214.395   & 1,608,041  & 43,019   \\
        \noalign{\smallskip}\hline
    \end{tabular}
\end{table*}

\subsubsection{Gold standard}
\label{subsec:goldstandard}

Assuming that we could find an ideal structure containing all interesting terms and relationships for each domain, the automatic evaluation task would become fairly easy -- we would only need to compare the relations extracted or the taxonomic structure with the gold standard, and the quality of the structure would be computed by the overlap between the two. 
Unfortunately, in practice, such gold standards do not exist. 
Thus, for the automatic evaluation, we employ the widely accepted, but not perfect, taxonomies as gold standards: 

\textbf{\nf{WordNet (EN)}}: Princeton Wordnet~ \cite{Fellbaum1998} is considered the standard model of a lexical ontology for the English language, combining the traditional lexicographic information with modern computation. 
Words in WordNet are divided into adjectives, adverbs, nouns, verbs and functional words. 
WordNet is structured into synsets, \idest~a set of synonyms that intends to represent the concept of its composed words. 
All synsets in WordNet are organized as a network of semantic relations (\exemp, hyponymy for nouns and troponymy for verbs). 
According to Tudhope \etal~\cite{TudhopeEtAl2006}, WordNet is the most widespread lexical database, containing 147,278 unique terms (nouns, verbs, adjectives, and adverbs) organized into 206,941 synsets\footnote{http://wordnet.princeton.edu/wordnet/man/wnstats.7WN.html}. 
Although WordNet is a widespread lexical database, it does not include much domain-specific terminology. 
In this work we use the version 3.0 of WordNet which is embedded into Natural Language Toolkit (NLTK)\footnote{http://www.nltk.org/}. 
NLTK provides a WordNet interface containing methods to access synsets, lemmas and relations such as hypernymy and hyponymy. 

\textbf{\nf{Onto.PT (PT)}}: Onto.PT\footnote{http://ontopt.dei.uc.pt/}~\cite{Oliveira2013,Oliveira2014} is a lexical ontology for Portuguese, structured in a similar fashion to WordNet, but differently than WordNet it was not manually built. 
As WordNet, it also groups synonymous words in synsets which are lexicalizations of a concept, and semantic relations are held between synsets. 
In order to build a system capable of automatically creating a semantic knowledge resource from other available resources, Oliveira~\cite{Oliveira2013,Oliveira2014} takes advantage of available NLP tools.  Onto.PT integrates the lexical-semantic network PAPEL \cite{OliveiraEtAl2009}, the electronic dictionaries Dicion\'ario Aberto \cite{SimoesEtAl2012} and Wiktionary.PT\footnote{http://pt.wiktionary.org/}, and three public synset-based thesauri namely: TeP 2.0 \cite{MazieroEtAl2008}, OpenWordNet-PT\footnote{https://github.com/arademaker/wordnet-br} and OpenThesaurus.PT\footnote{http://openthesaurus.caixamagica.pt/}. 
In this work we use the version 0.6 of Onto.PT in a WordNet RDF/OWL Basic\footnote{http://www.w3.org/2006/03/wn/wn20/} model which contains 67,873 instances of NomeSynset (noun-based synsets), 26,451 instances of VerboSynset (verb-based synsets), 20,760 instances of AdjectivoSynset (adjective-based synsets) and 2,366 instances of AdverbioSynset (adverb-based synsets). 
Synsets in the lexical ontology are connected in a network containing 341,506 relations in which 79,425 belong to the type \nf{hiperonimoDe} (hypernym-of). 
Other relations in Onto.PT include \nf{parteDe} (part-of), \nf{membroDe} (member-of) and \nf{temQualidade} (has-quality). 

\begin{figure}[t!]
    \centering
    \includegraphics[width=0.85\textwidth]{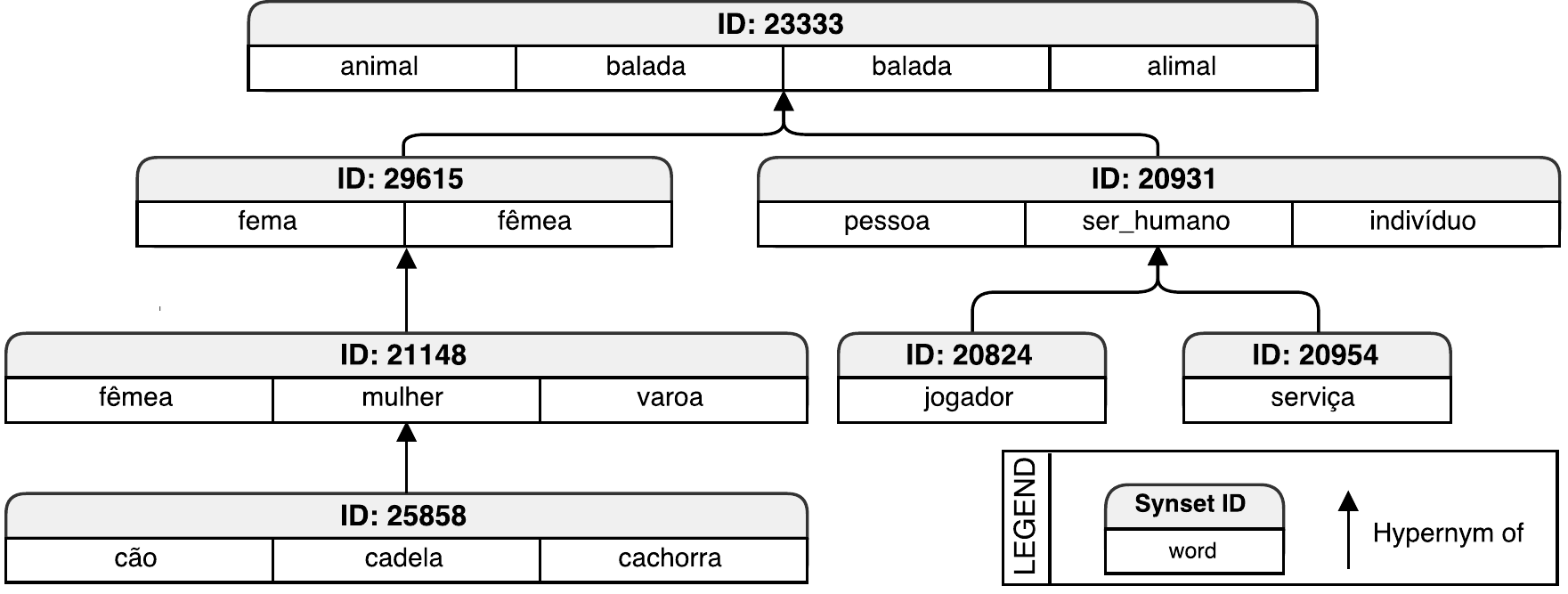}
    \caption{Excerpt of the extracted structure from Onto.PT.}
    \label{fig:excerptontopt}
\end{figure}

In the lack of an interface for the Onto.PT RDF/OWL file, we identified all synsets marked in RDF as \nf{NomeSynset} meaning that all our taxonomic relations occur only between nouns. 
For each identified synset we extracted its id, its lemmas (marked in RDF as \nf{formaLexical}) and its taxonomic relations (marked in RDF as \nf{hiperonimoDe} and \nf{hiponimoDe}). 
A directed graph (digraph) is created where each synset is a node and each taxonomic relation is an edge that connects two nodes. 
The digraph allows an easier access to hypernyms in higher levels such as the hypernym of a hypernym. 
An excerpt of the generated structure can be seen in Figure \ref{fig:excerptontopt}.

As Onto.PT is constructed grouping resources automatically, we can observe that there are some errors such as the word ``f\^emea'' (female) which repeats in two synsets (ID: 21615 and ID: 21148). 
In these cases we eliminate self-cycles, \idest, a term that is hypernym of itself. 
As this resource is automatically generated, it also may contain errors, such as the word ``mulher'' (womam) ID: 21148 which should not an hypernym of ``c\~ao'' (dog). 

\subsection{Pre-Processing}
\label{subsec:preprocessing}

Pre-processing is divided into parsing plain text files and how the context of each term is extracted. 

\subsubsection{Parsing}
\label{subsec:parsing}

Term extraction is an important part of taxonomy extraction and ontology construction \cite{LopesVieira2012}, since it extracts relevant terms of the domain. Usually term extraction requires a pre-processing in order to identify Part-of-Speech (PoS) tags or relations between words in a syntactic level (chunking). 
In this work we use  syntactic parsers to extract noun-phrases. 
The English corpora were parsed using the Stanford Lexicalized Parser \cite{KleinAndManning2003} (version 3.3.1), a well known parser and widely used in relation extraction \cite{LiEtAl2014,MeijerEtAl2014,XuEtAl2014}. 
For the Portuguese corpora we applied the PALAVRAS \cite{Bick2000} parser, which has been used in many work \cite{Beck2011,LopesVieira2012,AraujoEtAl2013,TabaCaseli2014}. 

\subsubsection{Context extraction}
\label{subsec:context_extraction}

Context extraction is used to select terms and contexts for the automatic evaluation process. 
As our evaluation is performed using WordNet \cite{Fellbaum1998} and Onto.PT \cite{Oliveira2013,Oliveira2014} as gold standards, all models for the automatic evaluation use single nouns instead of noun phrases, since the number of noun phrases in both resources is very limited. 
Also, WordNet can not distinguish that ``small dog'' is a ``dog'' modified by an adjective, and thus ``small dog'' is a ``dog'' or that ``small dog'' is a kind of ``animal'', as well as Onto.PT can not understand that ``cachorro pequeno'' is the noun ``cachorro''  modified by an adjective. 

\begin{figure}[tbh]
    \centering
    \includegraphics[width=0.85\textwidth]{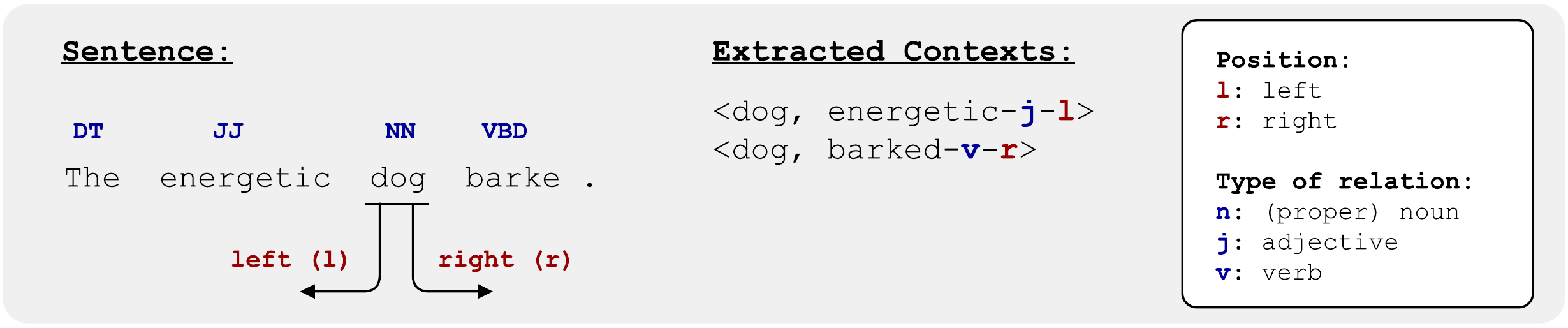}
    \caption{Example of context extraction of a sentence.}
    \label{fig:context}
\end{figure}

The first step of the context extraction collects all co-occurrences between a target word (a noun or a proper noun in the corpus) and content words, \idest, nouns, proper nouns, verbs and adjectives. 
The co-occurrences are extracted by sliding a window with a pre-defined length $n$ along the text. 
The size of the window was set to 5, \idest, every content word that occurs within a window of two words before or after is counted as a co-occurrence for the target word. 
Each context is marked with its relative position in relation to the target word. 
For example, consider the phrase ``The energetic dog barked.'' and the relations defined as \nf{$<$target word, context$>$} in the extraction presented in Figure \ref{fig:context}.

The relations \nf{$<$dog, energetic-j-l$>$} and \nf{$<$dog, barked-v-r$>$} are extracted, where the two characters at the end of the context means that the term contains determined Part-of-Speech (PoS) tag and is placed at determined position of the target noun. 
Thus, \nf{v-r} in \nf{$<$dog, barked-v-r$>$} means that the verb ``barked'' is placed on the right of the target noun ``dog''. The outcome of the this process is a list containing nouns and their contexts followed by the frequency of these contexts.

For models that use the distribution of words across documents instead of a window of size=5, the process of extracting contexts is different. 
For these models, all nouns and pronouns are extracted and their context is determined as the name of the file where they were extracted from. 
Thus, considering a file named ``doc\_1.txt'' containing the words ``dog'' and ``cat'' and another file named ``doc\_2.txt'' containing the words ``dog'' and ``fish'', the context of each word is: \nf{$<$dog, doc\_1.txt$>$}, \nf{$<$cat, doc\_1.txt$>$}, \nf{$<$dog, doc\_2.txt$>$} and \nf{$<$fish, doc\_2.txt$>$}. 
The outcome of the this process is a list containing nouns and their contexts followed by the frequency of the word in each contexts. 

As generating models containing all terms is a time and resource consuming task, we decided to reduce our list of target words. 
To reduce our list we took into account only target words that appear into the gold standard and share the highest number of contexts. 
Thus, for each target word existing in the corpus and in the gold standard we counted the number of contexts and selected the top $n$ target words. 
The evaluation was performed using $n$=1,000 and $n$=10,000.

\subsection{Relation extraction methods implementation}
\label{subsec:models}

For the automatic evaluation we developed seven methods for English and Portuguese languages. 
The unique method that cannot use the same implementation in both languages is the pattern-based method (\nf{Patt}) since it is a language-oriented model. 
For that method we have one implementation for English and another for Portuguese. 
The other methods use the same implementation for both languages, changing only the process of parsing. 
After parsing, lists are created containing nouns and their contexts. 
Having these lists, we built the following models: 

\textbf{\nf{DSim}}: The model based on Directional Similarity takes into account the Distributional Inclusion Hypothesis, according to which the contexts of a narrow term are also shared by the broad term (see Section \ref{subsec:distributional_inclusion}). 
This model uses the list of terms and contexts extracted using a window of size=5. 
The degree of association between terms and contexts is determined by a weight function. 
Thus, the value of the frequency of a term with a context is replaced by its Positive Pointwise Mutual Information (PPMI) \cite{ChurchHanks1990} value, where all negative values are set to zero. 
For computing the directional similarity we tested the measure proposed by Weeds \etal~\cite{WeedsEtAl2004} and the measure proposed by Clarke~\cite{Clarke2009} (hereafter \nf{ClarkeDE}). 
We decided to use these two directional similarity measures among all the existent because the taxonomic relation can be inferred by its precision and recall instead of defining a threshold. 
As the results using both measures were almost the same in almost all corpora, we decided to use in the evaluation process only the values generated by \nf{ClarkeDE} measure. 
The code for implementation these measures are freely available by Weeds\footnote{https://github.com/SussexCompSem/learninghypernyms}~\cite{WeedsEtAl2014}. 

\textbf{\nf{SLQS}}: The model based on entropy was developed by Santus \etal~\cite{SantusEtAl2014} and relies on the idea that superordinare terms are less informative than their hyponyms. 
This model also uses the list of terms and contexts extracted using a window of size=5. 
The difference when compared with \nf{DSim} model is that \nf{SLQS} model employs Local Mutual Information (LMI) to weight co-occurrences as well as uses entropy as an estimate of context informativeness. 
After extracting co-occurrences and weighting them using LMI, the $N$ most associated contexts are identified using the Shannon entropy measure \cite{Shannon1948}, where $N$ is set to 50. 
The resulting values of entropy are normalized using the Min-Max-Scaling in a range 0--1. 
Finally, the entropy of a word is defined as the median entropy of its $N$ contexts.

\textbf{\nf{TF}}: The model based on the frequency takes into account the number of times a word occur in the whole collection as an indicative of generalization-specialization. 
The idea in this model is that the more general a word, the higher its frequency. 
This model uses the list of terms extracted using documents as contexts. 
For each word, the resulting frequency is the sum of all individual frequencies in documents.

\textbf{\nf{DocSub}}: The model based on document subsumption uses the probability of the distribution of the words across shared documents in order to identify a taxonomic relation between them. 
According to this model, a word that appears in more documents tends to be more general than a word that appears in a subset of these documents. 
It is important to note that in this model a word subsumes another word only if it appears in a subset of the documents that the other word appears. 
A threshold indicating the percentage of shared documents may be set. 
This model uses the list of terms extracted using documents as contexts, where we can verify the intersection of documents shared by two words. 

\textbf{\nf{DF}}: The model based on the document frequency takes into account the number of documents in which a word appear as an evidence of taxonomic relation. 
Thus, a word that occur in more documents tends to be more general that a word that appear in few documents. 
This model is different than \nf{DocSub} because it takes into account only the number of documents in which a word appear and not the number of shared documents. 
This model uses the list of terms extracted using documents as contexts, where the frequency is represented by the number of documents in which a word occurs.

\textbf{\nf{HClust}}: The model based on hierarchical clustering uses contexts to group similar words together and the document frequency to identify the taxonomic relations between them. 
This model uses both lists of terms extracted using a window of size=5 and documents as contexts. 
The first list is used to hierarchically cluster similar words together, \idest, words that share similar contexts. 
The second list is used to identify the taxonomic relation between terms based on the document frequency as it occurs in \nf{DF} model. 
The difference of this model when compared with \nf{DF} is that the former refines the latter verifying the whether the taxonomy exists only for semantically related terms. 
The bottom up clustering, grouping the most similar terms together is performed using the freely available scripts of Fastcluster\footnote{http://danifold.net/fastcluster.html}. 

\textbf{\nf{Patt}}: The model based on patterns extracts taxonomic relations between words using rules developed by Hearst \cite{Hearst1992,Hearst1998} and its adapted rules for Portuguese \cite{Basegio2007}. 
This model is the unique language dependent and different rules run on English and Portuguese. 
We performed to approaches to extract nouns. 
The first one extract patterns using NPs and the comparison against the gold standard uses the head of the NP. 
The second approach extracts patterns based on nouns and thus, discarding relations when a pos-adjective appears. 

\subsection{Evaluation}
\label{sec:evaluations}

This evaluation intends to automatically compare the relations extracted from each model with the entries in the gold standards. 
A gold standard structure is a taxonomy that serves as reference. 
Thus, the quality of the extracted relations are expressed by their similarity to the gold standard hierarchy. 
In this work we use WordNet \cite{Fellbaum1998} as gold standard for English and Onto.PT \cite{Oliveira2014} as gold standard for Portuguese. 

\begin{figure}[t!]
    \centering
    \includegraphics[width=0.7\textwidth]{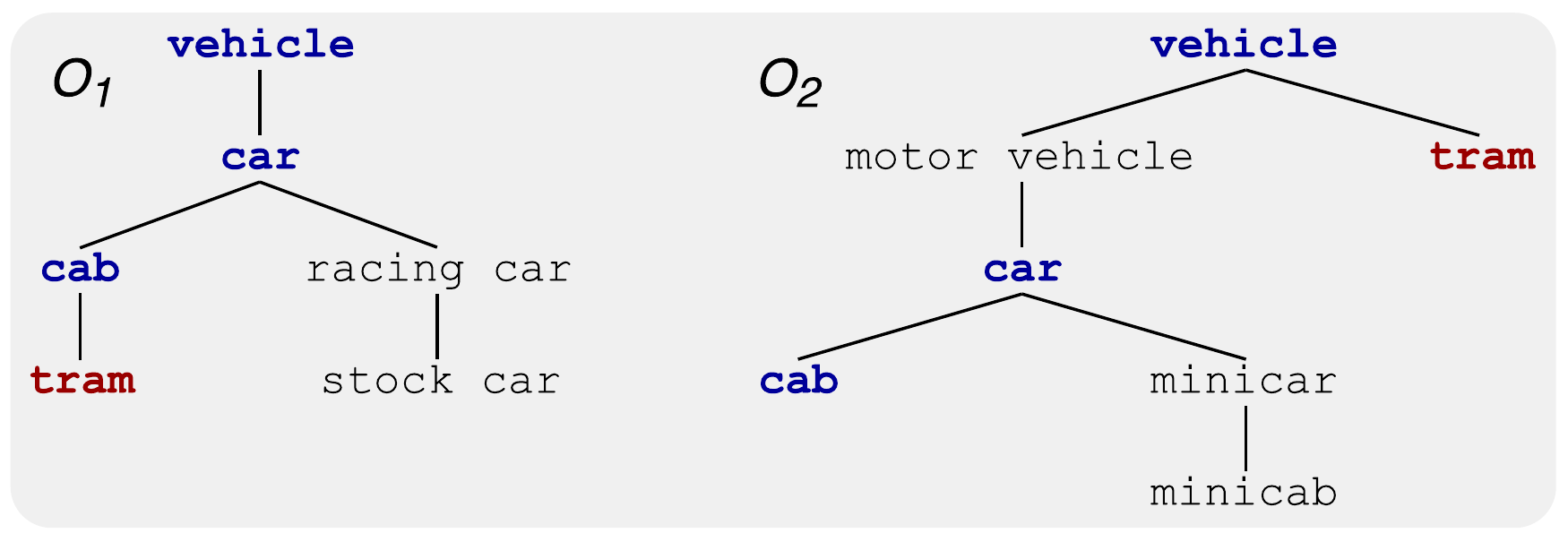}
    \caption{Taxonomies for common semantic cotopy example.}
    \label{fig:semantic_cotopy_example}
\end{figure}

The quality of the extracted relations may be measured in terms of Precision, Recall and F-measure when comparing with the gold standard. 
In order to achieve such measures, we first define the common relations (\nf{CR}) as the relations between a term and its super- and sub-terms that appear in both the extracted taxonomy and the gold standard:

\begin{equation}
CR(c,O_1, O_2) = \{c_i \in C_1 \cap C_2 |\: rel \} \ \ \ \ \ \ \ rel = \left\{\begin{matrix}
(c_i, c) & \text{if}\ c_i <_{c_1} c  \\ 
(c, c_i) & \text{if}\ c <_{c_1} c_i 
\end{matrix}\right.
\end{equation}

\noindent
where $O_1$ and $O_2$ are two taxonomies, $c$ is the term being analyzed, $c_i$ is a term common to both taxonomies, $C_1$ is the set of terms in $O_1$, $C_2$ is the set of terms in $O_2$, and $<_{c_1}$ is the partial order induced by the relationship in $O_1$. 
Take for instance the taxonomies in Figure \ref{fig:semantic_cotopy_example}, assuming that the left taxonomy ($O_1$) was extracted by some method and the taxonomy on the right ($O_2$) is the gold standard, the common relations for the term ``car'' in the taxonomy $O_1$ are \{(``vehicle'', ``car''), (``car'', ``cab''), (``car'', ``tram'')\}. 
As \nf{CR} only takes into account terms shared by both taxonomies, the set of relations for ``car'' also contains the relation (``car'', ``tram'') even if this relation does not exist in the gold standard. 
It is also important to note that even if the gold standard does not contain the direct relation between ``car'' and ``vehicle'', this relation is inherited by transitivity. 

Using the common relations (\nf{CR}), we can define precision ($\mathcal{P}$) and recall ($\mathcal{R}$) as:

\begin{equation}
\label{eqn:precision}
\mathcal{P} = \frac{\sum_{c \in C_T \cap C_{GS}} |CR(c, O_T, O_{GS}) \cap CR(c, O_{GS}, O_{T})|}{\sum_{c \in C_T \cap C_{GS}} |CR(c, O_T, O_{GS})|}
\end{equation}

\begin{equation}
\label{eqn:recall}
\mathcal{R} = \frac{\sum_{c \in C_T \cap C_{GS}} |CR(c, O_T, O_{GS}) \cap CR(c, O_{GS}, O_{T})|}{\sum_{c \in C_T \cap C_{GS}} |CR(c, O_{GS}, O_{T})|}
\end{equation}

\noindent
where $c$ is the term being analyzed, $C_T$ is the set of terms from the generated taxonomy, $C_{GS}$ is the set of terms from the gold standard, $O_T$ is the generated taxonomy, and $O_{GS}$ is the gold standard. 
Thus, the precision score is the result of dividing the amount of relations identified by the method that are in the gold standard by the total number of relations identified by the method. 
Recall is the score achieved by dividing the number of relations identified by the method and that are present in the gold standard by the total number of relations in the gold standard. 
F-measure is the score representing a weighted average of the precision and recall: 

\begin{equation}
\label{eqn:fmeasure}
\mathcal{F} = \frac{2 \times \mathcal{P} \times \mathcal{R} }{\mathcal{P} +  \mathcal{R}}
\end{equation}

\subsubsection{Metrics for characterizing taxonomies}
\label{subsec:metrics}

According to Vrandecic and Sure \cite{VrandecicSure2007}, measuring ontologies is necessary to evaluate them both during engineering and application, being a necessary precondition to perform quality assurance and control the process of improvement. 
Metrics allow the fast and simple assessment of an ontology and also to track their subsequent evolution. 
Indeed, they are expected to give some insight for ontology developers to help them to design ontologies, improve ontology quality, anticipate and reduce future maintenance requirements, as well as help ontology users to choose the ontologies that best meet their needs \cite{YaoEtAl2005}. 
A compilation of metrics for ontology evaluation is performed by Freitas and Vieira~\cite{Freitas2010,FreitasVieira2010}.

Although these metrics are applied to ontologies, they also can be applied to other structures such as taxonomies. 
In this work, besides interpreting methods in terms of automatic and manual evaluations, we also intend to describe methods in terms of the structure of the generated taxonomy. 
Thus, we analyze the subset of the metrics that are applied to taxonomies or Rooted Directed Acyclic Graphs (Rooted DAG), \idest, a structure having a single highest node (Root) and all other nodes are connected by means of \nf{is-a} links, generating a chain of links to the Root. 

\begin{figure}[tbh]
    \centering
    \includegraphics[width=0.7\textwidth]{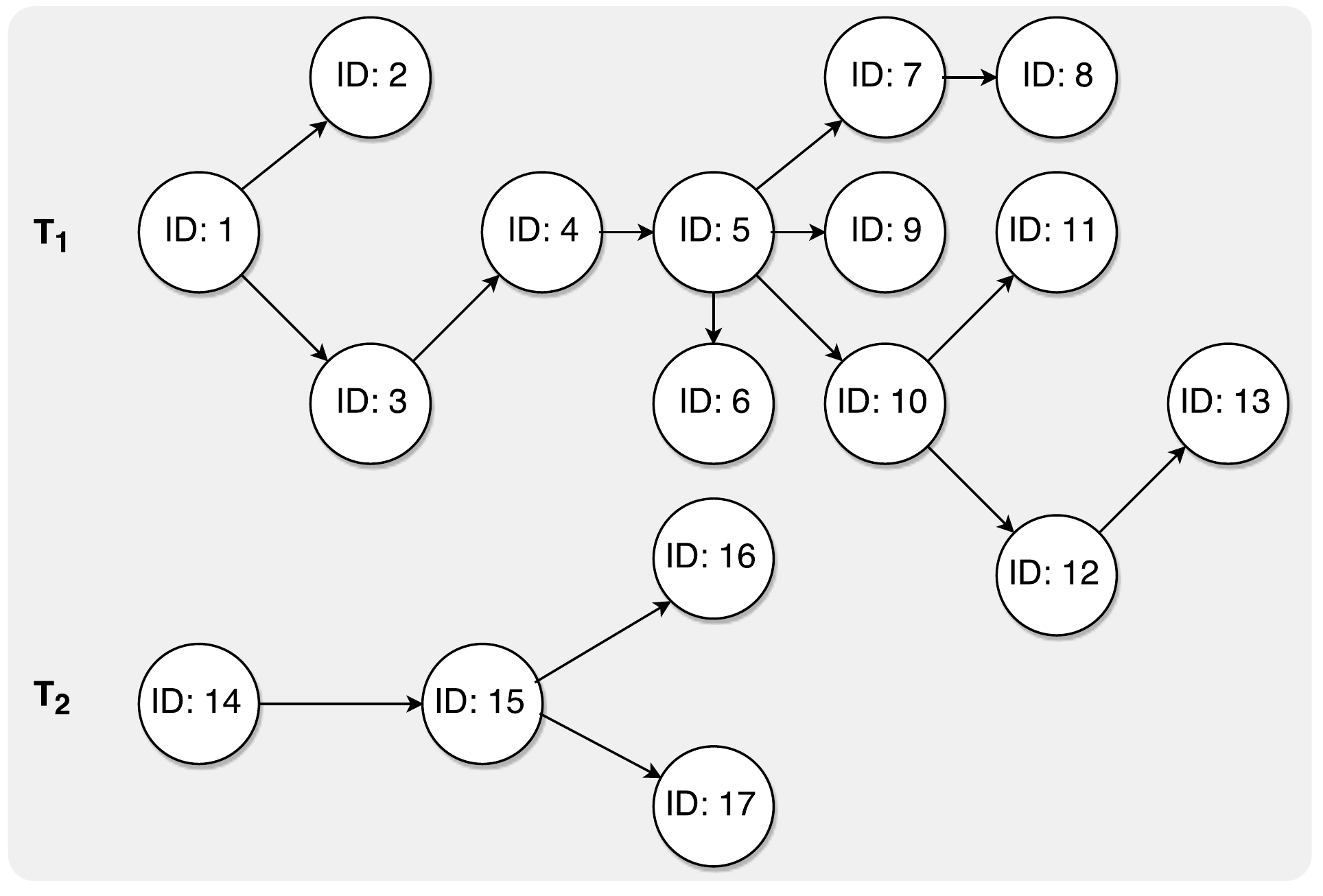}
    \caption{Examples of taxonomies represented as direct graphs.}
    \label{fig:taxonomy_example}
\end{figure}

In order to better understand the following metrics, consider that each applied method may generate one or more taxonomies, and each taxonomy has a root term and at least one leaf term. 
Examples presented in some metrics use Figure \ref{fig:taxonomy_example} where two taxonomies are represented as directed graphs (digraphs). 
Nodes represent words being identified by their ids, and edges represent the relation ``is hypernym of''. 
Thus, the connection between nodes ``ID: 1'' and ``ID: 2'' means that the word identified by ``ID: 1'' is hypernym of the word identified by ``ID: 2''. \

A set containing the main metrics are described below, where ``\nf{terms}'' refers to a set of terms, ``\nf{term\txtsub{ij}}'' refers to the $i$th term in the taxonomy $j$ and ``\nf{term\txtsub{ij}}'' $\in$ ``\nf{terms}'', ``\nf{Count()}'' is a function that counts the number of occurrences, ``\nf{Max()}'' determines the maximum value of the set, ``\nf{Min()}'' determines the minimum value of the set, ``\nf{isRoot()}'' and ``\nf{isLeaf()}'' verify whether the argument is a root or a leaf term in the taxonomy, ``\nf{Depth()}'' returns the depth of a term, ``\nf{Width()}'' returns the number of siblings of a term, and ``\nf{hasSiblings()}'' returns terms that have siblings. 

\vspace{5mm}
\noindent
\textbf{TotalTerms}: Total number of terms takes into account all unique terms generated by all taxonomies using a specific method. 
Example: The total number of terms in Figure \ref{fig:taxonomy_example} is: 17 (ID: 1 to ID: 17).
\begin{equation*}
 \mathtt{TotalTerms} = Count(\mathtt{terms})
\end{equation*}

\noindent    
\textbf{TotalRoots}: Total number of roots indicates the number of upper terms of all taxonomies, \idest, terms without hypernyms in a taxonomy. 
This means also the number of taxonomies generated by the method. 
Example: The total number of roots in Figure \ref{fig:taxonomy_example} is: 2 (ID: 1 and ID: 14).
\begin{equation*}
 \mathtt{RootTerms} = Count(isRoot(\mathtt{term_{ij}}))
\end{equation*}

\noindent
\textbf{NumberRels}: Number of relations extracted from the corpus without count repetitions.
\begin{equation*}
 \mathtt{NumberRels} = Count(noRepetition(\mathtt{rel_{ij}}))
\end{equation*}

\noindent
\textbf{MaxDepth}: Maximum depth extracts the longest path between a root and a leaf for each taxonomy and select the maximum value. 
Example: The maximum depth in Figure \ref{fig:taxonomy_example} is: 6 (passing by IDs: 1, 3, 4, 5, 10, 12 and 13 in T1)
\begin{equation*}
 \mathtt{MaxDepth} = Max(Depth(isLeaf(\mathtt{term_{ij}})))
\end{equation*}

\noindent
\textbf{MinDepth}: Minimum depth extracts the shortest path between a root and a leaf for each taxonomy and select the minimum value. 
Example: The minimum depth in Figure \ref{fig:taxonomy_example} is: 1 (path between ID:1 and ID:2 in T1)
\begin{equation*}
 \mathtt{MinDepth} = Min(Depth(isLeaf(\mathtt{term_{ij}})))
\end{equation*}

\noindent
\textbf{AvgDepth}: Average depth is the ratio between the sum of all depths and the total number of taxonomies. 
Example: The average depth in Figure \ref{fig:taxonomy_example} is: (1+4+5+4+5+6+2+2)/2 = 14.5
\begin{equation*}
 \mathtt{AvgDepth} = \frac{\sum_{j} Depth(isLeaf(\mathtt{term_{ij}}))}{\mathtt{TotalRoots}}
\end{equation*}

\noindent
\textbf{DepthCoesion}: The coesion of a taxonomy is indicated by the maximum depth divided by its average depth. 
Example: The coesion of the taxonomy in Figure \ref{fig:taxonomy_example} is: (6/14.5) $\approx$ 0.41
\begin{equation*}
 \mathtt{DepthCoesion = \frac{MaxDepth}{AvgDepth}}
\end{equation*}

\noindent
\textbf{MaxWidth}: Maximum width is the maximum number of term siblings in all taxonomies, \idest, the maximum number of hyponyms of a term. 
Example: The maximum width in Figure \ref{fig:taxonomy_example} is: 4 (formed by IDs: 6, 7, 9 and 10 in T1)
\begin{equation*}
 \mathtt{MaxWidth} = Max(Width(\mathtt{term_{ij}}))
\end{equation*}

\noindent
\textbf{MinWidth}: Minimum width is the minimum number of term siblings in all taxonomies, \idest, the minimum number of hyponyms of a term. 
Example: The minimum width in Figure \ref{fig:taxonomy_example} is: 1 (single IDs: 4, 5, 8, 13, or 15)
\begin{equation*}
 \mathtt{MinWidth} = Min(Width(\mathtt{term_{ij}}))
\end{equation*}

\noindent
\textbf{AvgWidth}: Average width is the ratio between the sum of widths (\idest, the sum of hyponyms) and the total number of siblings (\idest, the total number of terms that have hyponyms), and the number of taxonomies, where ``TaxWidth$_j$'' measures the width of the taxonomy $j$. 
Example: The average width in Figure \ref{fig:taxonomy_example} is: (((2+1+1+4+1+2+1)/7) + ((1+2)/2))/2 $\approx$ 1.61
\begin{equation*}
 \mathtt{TaxWidth_j} = \frac{\sum_{i}  Width(\mathtt{term_{ij}}))}{Count(hasSiblings(\mathtt{term_{ij}}))}
\end{equation*}
\begin{equation*}
 \mathtt{AvgWidth} = \frac{\sum_{j} \mathtt{TaxWidth_j}}{\mathtt{TotalRoots}}
\end{equation*}

\subsubsection{Complementarity Analysis}

It is interesting to observe all generated taxonomies, and to analyze how models are complementary, \idest, verify whether the same relations are generated by more than one model. 
It is also interesting to observe whether relations are generated in opposite directions (\exemp, \nf{Patt} model generates the relation \nf{A}$\rightarrow$\nf{B} and \nf{DSim} model generates the inverse relation \nf{B}$\rightarrow$\nf{A}). 
To illustrate the effect of mixing models, we use color maps with the ratios of direct relations, \idest~relations that are equal in both models, and inverse relations. 

\section{Evaluation of Taxonomy Extraction Methods}
\label{sec:evaluation}

This section presents a series of experiments aiming to evaluate the methods presented in Section \ref{subsec:models}. 
In order to verify the quality of the extracted relations and indirectly the quality of the method that generated such relations, we performed automatic evaluations based on the comparison of the relations extracted from each model with a gold standard, resulting in precision, recall and f-measure for each method. 

The automatic evaluation is performed on relations extracted from Europarl and TED Talks corpora in English and Portuguese, using 7 models: \nf{Patt}, \nf{DSim}, \nf{SLQS}, \nf{TF}, \nf{DF}, \nf{DocSub} and \nf{HClust}. 
As \nf{DocSub} model may generate different values of precision, recall and f-measure according to the threshold, a deeper analysis on the results generated is performed, as well as for \nf{HClust} model, since this model can generate different results according to the selected number of clusters. 
Section \ref{subsec:metrics_analysis} presents an analysis on the characteristics of the taxonomies generated by each method. 
These characteristics include the number of generated relations, number of taxonomies, depth of the taxonomy, \etc. 
Finally, we analyze the complementarity of the developed models in Section \ref{subsec:complementarity}.

\subsection{Automatic quality analysis}

Zipf's law~\cite{Zipf1935} states that the relationship between a word's frequency and the rank order of its frequency is roughly a reciprocal curve, \idest, if we count up how often each word occurs in a large corpus, and then sort the words by their frequency of occurrence, there exists a relationship between the frequency of a word and its position in the sorted list (called rank). 
For instance, according to this law, the 50th most common word in the corpus should occur with three times the frequency of the 150th most common word. 

Because of this Zipfian distribution of words, cutting out low frequency words will greatly reduce our space (as well as the memory requirements of the system), while not considerably affecting the model quality. 
Hence, we decided to reduce our vocabulary. 
The first experiment reduces the vocabulary to 1,000 terms, a second experiment considers 10,000 terms and a third experiment considers 1,000 terms but applying a filtering algorithm to induce a taxonomy where each term contains only one hypernym. 
As in our experiments the word frequency is directly related to the number of contexts, due to the fact that words are extracted in a window, our vocabulary was reduced taking into account only terms that share the highest number of contexts and appear into the gold standard. 
Thus, for each word existing in the corpus and in the gold standard we counted the number of contexts and selected the top $N$ terms. 
Using the new vocabulary we applied all models described in Section \ref{subsec:models} and compared the result of each method with the gold standard, generating values of precision ($\mathcal{P}$), recall ($\mathcal{R}$) and f-measure ($\mathcal{F}$) for each model in each corpus. 

\begin{itemize}
\item {\bf Experiment 1: Using 1,000 terms}
\end{itemize}

In this first experiment we reduce the vocabulary to the top 1,000 terms with the highest number of contexts. 
Table \ref{tab:top1000} presents the general overview of these values for each method in each corpus, where the highest values are presented in bold. 
As \nf{DocSub} and \nf{HClust} can generate a range of values of precision and recall according to the threshold, the table contains the values with the highest value of f-measure (\nf{DocSub}: $\lambda$=0.1 using all corpora but TED Talks in English which has $\lambda$=0.3, \nf{HClust}: 1,000 clusters for all corpora). 
\nf{Patt} contains values of precision and recall and f-measure considering all rules for the limited vocabulary.

\begin{table*}[htbp]
 \caption{Precision, recall and F-measure for methods using the top 1,000 words with the highest number of contexts.}
 \label{tab:top1000}
 \centering
 \begin{tabular}{c|cl|ccccccc}
   \hline\noalign{\smallskip}
    \noalign{\smallskip}
                                    & Lang           & Corpus    & Patt        & DSim   & SLQS   & TF          &   DF        & DocSub      & HClust \\
    \noalign{\smallskip}\hline\noalign{\smallskip}
    \multirow{4}{*}{$\mathcal{P}$}  & \multirow{2}{*}{EN} & Europarl  &{\bf 0.1173} & 0.0366 & 0.0503 & 0.0554      & 0.0548      & 0.0443      & 0.0761 \\
                                    &                     & Ted Talks &{\bf 0.1125} & 0.0301 & 0.0382 & 0.0425      & 0.0441      & 0.0710      & 0.0664 \\
                        \cline{2-10}\noalign{\smallskip}
                                    & \multirow{2}{*}{PT} & Europarl  & 0.5163      & 0.3330 & 0.5257 & 0.6109      & 0.5984      &{\bf 0.7311} & 0.5676 \\
                                    &                     & Ted Talks & 0.5387      & 0.2907 & 0.5300 & 0.6117      & 0.6159      &{\bf 0.6533} & 0.5656 \\
    \noalign{\smallskip}\hline\hline\noalign{\smallskip}
    \multirow{4}{*}{$\mathcal{R}$}  & \multirow{2}{*}{EN} & Europarl  & 0.0396      & 0.3999 & 0.5499 &{\bf 0.6045} & 0.5887      & 0.0023      & 0.0017 \\
                                    &                     & Ted Talks & 0.0018      & 0.4442 & 0.5377 & 0.5657      &{\bf 0.6077} & 0.2666      & 0.0019 \\
                        \cline{2-10}\noalign{\smallskip}
                                    & \multirow{2}{*}{PT} & Europarl  & 0.0111      & 0.3554 & 0.5795 &{\bf 0.6727} & 0.5184      & 0.0053      & 0.0012 \\
                                    &                     & Ted Talks & 0.0004      & 0.3142 & 0.5484 &{\bf 0.6877} & 0.5515      & 0.4706      & 0.0011 \\
    \noalign{\smallskip}\hline\hline\noalign{\smallskip}
    \multirow{4}{*}{$\mathcal{F}$}  & \multirow{2}{*}{EN} & Europarl  & 0.0591      & 0.0671 & 0.0922 &{\bf 0.1015} & 0.1003      & 0.0044      & 0.0033 \\
                                    &                     & Ted Talks & 0.0035      & 0.0564 & 0.0713 & 0.0791      & 0.0822      &{\bf 0.1121} & 0.0037 \\
                        \cline{2-10}\noalign{\smallskip}
                                    & \multirow{2}{*}{PT} & Europarl  & 0.0217      & 0.3438 & 0.5513 &{\bf 0.6403} & 0.5555      & 0.0105      & 0.0024 \\
                                    &                     & Ted Talks & 0.0008      & 0.3020 & 0.5390 &{\bf 0.6475} & 0.5819      & 0.5471      & 0.0022 \\
 \noalign{\smallskip}\hline
 \end{tabular}
\end{table*}

Analyzing Table \ref{tab:top1000}, we can observe that all values of precision using the Portuguese corpora have higher scores when compared with the English corpora. 
This is due to the fact that the Portuguese gold standard, Onto.PT, has more connections between synsets. 
For example, the word ``dog'' in WordNet is present in a total of 7 synsets, its direct hypernyms (first level of hypernyms) are distributed into 8 synsets and the second level of hypernyms (hypernyms of hypernyms of the word ``dog'') contains 16 synsets. 
On the other hand, the word ``cachorro'' (dog) in Onto.PT appears in 4 synsets, its direct hypernyms are distributed into 21 synsets and the second level of hypernyms contains 33 synsets. 
A higher number of terms associated in hypernyms tends to increase the precision. 
Another aspect to be considered is the fact that as Onto.PT is automatically constructed, there are relations that would not exist if it was manually constructed or revised. 
For instance, a synset containing the word ``homem'' (man) is hypernym of a synset containing the words ``cara'' (face), ``face'' (face), ``fronte'' (forehead) and ``testa'' (forehead). 
Although the polyssemy of the word ``cara'' (face, but also guy) may make the taxonomic relation with ``homem'' correct, since ``cara'' (guy) is a kind of ``homem'' (man), the other words that belong to the same synset will make this relation wrong, since ``cara'' (face) is not a kind of ``homem'' (man). 
Assuming that the relation between both synsets is correct because we have a polyssemic word will make relations such as ``testa'' (forehead) is a kind of ``homem'' (man) also correct, even if they are not. 

As we can observe in Table \ref{tab:top1000}, \nf{Patt} has the best values of precision for the English corpora while \nf{DocSub} has the best values for the Portuguese corpora. 
\nf{TF} has the best values of recall and f-measure for all corpora but the English version of TED Talks which has in \nf{DF} the best value of recall and in \nf{DocSub} the best value of f-measure. 
It was expected quite similar values of precision, recall and f-measure between \nf{TF} and \nf{DF} using the Europarl corpora since the size of each document was set to the size of the phrase because Europarl does not have document borders. 
Thus, terms that appear only once in a phrase have the same value of \nf{TF} and \nf{DF}. 
This value differs when a term appears more than once in a phrase, and thus, having a higher value of \nf{TF} than \nf{DF}. 
In some cases it seems to make difference in results, \exemp, Europarl in Portuguese which increased the precision from $\mathcal{P}$=0.5984 in \nf{DF} to $\mathcal{P}$=0.6109 in \nf{TF}, as well as the recall from $\mathcal{R}$=0.5184 in \nf{DF} to $\mathcal{R}$=0.6727 in \nf{TF}, resulting in an increase of f-measure from $\mathcal{F}$=0.5555 in \nf{DF} to $\mathcal{F}$=0.6403 in \nf{TF}. 
On the other hand, TED Talks corpora, which have document borders and each document contains dozens phrases, have similar values when compared with Europarl corpora. 
It makes sense since we are using terms that share a the highest number of contexts and thus, terms that should appear in a greater number of documents.

When comparing \nf{DF} model which takes into account only the number of documents that the word occurs, with \nf{DocSub} which considers the number of shared documents between two words, \nf{DocSub} achieved better values of precision, but lower values of recall. 
In fact, \nf{DocSub} had worse results in precision only when using Europarl corpus in English, where \nf{DF} reached best values of precision and f-measure. 
As \nf{DocSub} uses the shared documents, it seems reasonable that it has lower recall when compared with \nf{DF}. 
A further analysis on the distribution of precision, recall and f-measure using the \nf{DocSub} model shows that the highest values of f-measure were obtained using very low thresholds (when the number of documents shared are close to 10\%), and thus, approximating to the values of \nf{DF}.

Another interesting observation is to compare the results obtained by \nf{DF} with the results achieved by \nf{HClust}. 
This comparison is interesting since \nf{HClust} uses the values of document frequency over semantically clustered terms. 
By clustering semantically related terms, the \nf{HClust} model intends to increase the precision of the extracted relations forgoing the recall. 
As we can observe, it seems that clustering semantically related terms will increase the precision (at least for the top 1,000 terms in the English corpora used in this experiment) as expected. 
On the other hand, the problem of clustering similar terms is that terms that occur in the same contexts tend to be synonyms or co-hyponyms instead of hypernyms. 
Thus, \nf{HClust} may cluster also synonyms or co-hyponyms that differ in terms of document frequency and thus, the model classify them as a hypernym-hyponym relation. 
For instance, the cluster containing the highest value of similarity using the TED Talks corpus in English is composed by the terms ``consumer'' which occurs in 196 documents, and ``employee'' which occurs in 149 documents. 
Thus, according \nf{HClust} model the relation $<$consumer, \nf{is-a}, employee$>$ holds. 
Analyzing their relation in WordNet, they both are hyponym of the synset ``person.n.01'' but there is not a taxonomic relation between them.

Low values of precision were expected for methods that use the distribution of the words across documents due to the fact that such methods are good to indicate a semantic relation between terms, but not really good to identify the type of semantic relation. 
On the other hand, values of precision were expected to be high for the method based on patterns (\nf{Patt}), since they are well known for having high precision and low recall values. 
In fact, compared to the other methods, \nf{Patt} achieved better precision and lower recall, although it is still a low precision when compared with other work. 
\nf{DSim} obtained the lowest scores of precision, recall and f-measure, lower than the usual baseline \nf{TF}. 
This low result might be because many words do not share any context with other words and, even if their relation exists in WordNet, it can not be detected by \nf{ClarkeDE} measure. 
As \nf{SLQS} takes into account the median entropy of all its most related contexts as an estimate of word informativeness, it does not need to share any context with the other word to identify their taxonomic relation. 
Also, it might work well in small data sets, since it uses a limited number of contexts to generate the median entropy. 
In our work, we used the same number of most associated contexts as the original work ($N=50$), being the contexts ranked by their value of Local Mutual Information (LMI) with the target word.

\begin{itemize}
\item {\bf Experiment 2: Using 10,000 terms}
\end{itemize}

Observing results from \nf{HClust}, which obtained the best f-measure in a cluster containing 1,000 terms (the f-measure was still rising), we decide to perform a second experiment, increasing the number of terms, and consequently clusters. 
Adding more terms and clusters could increase the number of semantic relations and maybe taxonomic relations would be more evident. 
As the number of terms may influence the results we decided to perform the experiment using up to 10,000 words in the dictionary. 
This number of words is higher than the maximum number of words in the Portuguese version of TED Talks after filtering terms with Onto.PT (7,066 words). 
The other corpora have more terms than this threshold (TED Talks in English contains 19,601 words, Europarl in Portuguese 13,139 words and Europarl in English 32,007 words). 
Table \ref{tab:top10k} presents the values of precision and recall for all models using a vocabulary containing up to 10,000 words, where \nf{DocSub} and \nf{HClust} contain results when the best f-measure was achieved, and \nf{Patt} consider all patterns with the limited number of words. 

\begin{table*}[ht]
 \caption{Precision, recall and F-measure for methods using the top 10,000 words with the highest number of contexts.}
 \label{tab:top10k}
 \centering
 \begin{tabular}{c|cl|ccccccc}
   \hline\noalign{\smallskip}
    \noalign{\smallskip}
                                    & Lang            & Corpus    & Patt          & DSim   & SLQS   & TF          &   DF        & DocSub      & HClust \\
    \noalign{\smallskip}\hline\noalign{\smallskip}
    \multirow{4}{*}{$\mathcal{P}$}  & \multirow{2}{*}{EN} & Europarl  &{\bf 0.1192}   & 0.0083 & 0.0137 & 0.0150      & 0.0150      & 0.0445      & 0.0326 \\
                                    &                     & Ted Talks &{\bf 0.1022}   & 0.0069 & 0.0060 & 0.0092      & 0.0090      & 0.0356      & 0.0162 \\
                        \cline{2-10}\noalign{\smallskip}
                                    & \multirow{2}{*}{PT} & Europarl  & 0.5710        & 0.1948 & 0.3855 & 0.5474      & 0.4485      &{\bf 0.8052} & 0.4058 \\
                                    &                     & Ted Talks &{\bf 0.6304}   & 0.1870 & 0.3250 & 0.5312      & 0.4576      & 0.6064      & 0.3698 \\
    \noalign{\smallskip}\hline\hline\noalign{\smallskip}
    \multirow{4}{*}{$\mathcal{R}$}  & \multirow{2}{*}{EN} & Europarl  & 0.0037        & 0.3278 & 0.5941 & 0.6486      &{\bf 0.6490} & 0.0017      & 0.0003 \\
                                    &                     & Ted Talks & 0.0002        & 0.1486 & 0.4332 &{\bf 0.6467} & 0.6332      & 0.0967      & 0.0003 \\
                        \cline{2-10}\noalign{\smallskip}
                                    & \multirow{2}{*}{PT} & Europarl  & 0.0002        & 0.1562 & 0.5157 &{\bf 0.7255} & 0.5932      & 0.0032      & 0.0001 \\
                                    &                     & Ted Talks &2.10\txtsob{-5}& 0.0507 & 0.4492 &{\bf 0.7000} & 0.5887      & 0.1390      & 0.0002 \\
    \noalign{\smallskip}\hline\hline\noalign{\smallskip}
    \multirow{4}{*}{$\mathcal{F}$}  & \multirow{2}{*}{EN} & Europarl  & 0.0073        & 0.0162 & 0.0268 &{\bf 0.0293} &{\bf 0.0293} & 0.0033      & 0.0006 \\
                                    &                     & Ted Talks & 0.0004        & 0.0132 & 0.0118 & 0.0181      & 0.0179      &{\bf 0.0520} & 0.0005 \\
                        \cline{2-10}\noalign{\smallskip}
                                    & \multirow{2}{*}{PT} & Europarl  & 0.0005        & 0.1733 & 0.4412 &{\bf 0.6240} & 0.5109      & 0.0064      & 0.0002 \\
                                    &                     & Ted Talks &4.10\txtsob{-5}& 0.0798 & 0.3771 &{\bf 0.6040} & 0.5149      & 0.2261      & 0.0004 \\
 \noalign{\smallskip}\hline
 \end{tabular}
\end{table*}

As we can see, values of precision were lower for most methods, with exception of \nf{Patt} and \nf{DocSub}, which increased for most corpora. 
When increasing the number of terms to 10,000, the \nf{DocSub} models using Europarl corpora performed better than when using TED Talks corpora. 
It seems that the higher the number of documents, the more accurate relations become. 
Although decreasing the values of precision, \nf{TF} and \nf{DF} increased the values of recall, but decreasing the values of f-measure. 
As occurred in the experiment using the top 1,000 words, this experiment also kept \nf{TF} with the highest values of f-measure for most methods. 
\nf{TF} and \nf{DF} achieved almost the same values of precision, recall and f-measure using the English corpora, achieving the same value of precision ($\mathcal{P}$=0.0150) and f-measure ($\mathcal{F}$=0.0293) when using the Europarl corpus in English.

When comparing \nf{DF} with \nf{HClust}, it seems a good approach in English to verify the hierarchical relation only for terms that are semantically related instead of considering all terms. 
As occurred in the Experiment 1, \nf{DF} performed better than \nf{HClust} using the Portuguese corpora. 
The lowest values of precision are achieved by \nf{DSim} model, and the lowest recalls are obtained by \nf{HClust} and \nf{Patt} models. 
These models are known for having lower levels of recall since \nf{HClust} tries to increase precision in detriment of recall when clustering similar words, and \nf{Patt} model uses patterns that are very sparse in texts. 
As these models have very low values of recall, they also contain the lowest values of f-measure. 
Observing the increase of terms for hierarchical clustering, all values of precision, recall and f-measure decreased when comparing with the ones obtained using 1,000 terms. 

\begin{itemize}
\item {\bf Experiment 3: Selecting the best parent using 1,000 terms}
\end{itemize}

The third experiment intends to remove multiple hypernyms when it occurs to a term, maintaining the taxonomy with a tree structure. 
The decision for the correct hypernym of a term is based on a score calculated for each potential hypernym as described by De Knijff \etal~\cite{DeKnijffEtAl2013}. 
The score is defined in Equation \ref{eqn:best_parent} and takes into account the distance between the target term and the list of ancestors parents. 

\begin{equation}
 \label{eqn:best_parent}
 score(p,x) = P(p|x) + \sum_{a \in A_p}w(a,x) \cdot P(a|x)
\end{equation}

\noindent
where $p$ is the potential parent of term $x$ and $Ap$ is the list of ancestors of $p$. 
The co-occurrence probability $P(a|x)$ is multiplied by the weight $w(a,x)$. 
This weight is defined as:

\begin{equation}
w(a,x) = \frac{1}{d(a,x)}
\end{equation}

\noindent
where $d(a,x)$ is the path length between term $x$ and its ancestor $a$. 
After computing the $score(p,x)$ for all possible parents, the parent containing the highest score is chosen as parent of the term $x$. 

Table \ref{tab:top1000_1parent} presents the values of precision, recall and f-measure for the methods in all corpora. 
The filtering on multiple hypernyms is applied in relations extracted using 1,000 terms in the dictionary. 
Thus, Table \ref{tab:top1000} can be compared with Table \ref{tab:top1000_1parent}. 

\begin{table*}[ht]
 \caption{Precision, recall and F-measure for methods using the top 1,000 words with the highest number of contexts and selecting the best parent.}
 \label{tab:top1000_1parent}
 \centering
 \begin{tabular}{c|cl|ccccccc}
   \hline\noalign{\smallskip}
    \noalign{\smallskip}
                                    & Lang                & Corpus    & Patt        & DSim   & SLQS   & TF     &   DF   & DocSub      & HClust \\
    \noalign{\smallskip}\hline\noalign{\smallskip}
    \multirow{4}{*}{$\mathcal{P}$}  & \multirow{2}{*}{EN} & Europarl  &{\bf 0.1038} & 0.0170 & 0.0490 & 0.0641 & 0.0641 & 0.0613      & 0.0761 \\
                                    &                     & Ted Talks &{\bf 0.1282} & 0.0291 & 0.0410 & 0.0270 & 0.0270 & 0.1154      & 0.0661 \\
                        \cline{2-10}\noalign{\smallskip}
                                    & \multirow{2}{*}{PT} & Europarl  & 0.6185      & 0.3744 & 0.4144 & 0.4394 & 0.4394 &{\bf 0.7553} & 0.5676 \\
                                    &                     & Ted Talks & 0.6308      & 0.4124 & 0.4404 & 0.4515 & 0.4945 &{\bf 0.8609} & 0.5295 \\
    \noalign{\smallskip}\hline\hline\noalign{\smallskip}
    \multirow{4}{*}{$\mathcal{R}$}  & \multirow{2}{*}{EN} & Europarl  &{\bf 0.0021} & 0.0004 & 0.0011 & 0.0014 & 0.0014 & 0.0013      & 0.0017 \\
                                    &                     & Ted Talks & 0.0011      & 0.0008 & 0.0011 & 0.0008 & 0.0008 &{\bf 0.0030} & 0.0018 \\
                        \cline{2-10}\noalign{\smallskip}
                                    & \multirow{2}{*}{PT} & Europarl  & 0.0012      & 0.0008 & 0.0009 & 0.0010 & 0.0010 &{\bf 0.0016} & 0.0012 \\
                                    &                     & Ted Talks & 0.0003      & 0.0009 & 0.0009 & 0.0010 & 0.0010 &{\bf 0.0017} & 0.0011 \\
    \noalign{\smallskip}\hline\hline\noalign{\smallskip}
    \multirow{4}{*}{$\mathcal{F}$}  & \multirow{2}{*}{EN} & Europarl  &{\bf 0.0041} & 0.0007 & 0.0021 & 0.0027 & 0.0027 & 0.0026      & 0.0033 \\
                                    &                     & Ted Talks & 0.0022      & 0.0016 & 0.0022 & 0.0015 & 0.0015 &{\bf 0.0058} & 0.0036 \\
                        \cline{2-10}\noalign{\smallskip}
                                    & \multirow{2}{*}{PT} & Europarl  & 0.0024      & 0.0016 & 0.0018 & 0.0019 & 0.0019 &{\bf 0.0031} & 0.0023 \\
                                    &                     & Ted Talks & 0.0005      & 0.0018 & 0.0018 & 0.0020 & 0.0021 &{\bf 0.0034} & 0.0022 \\
 \noalign{\smallskip}\hline
 \end{tabular}
\end{table*}

Analyzing Table \ref{tab:top1000_1parent} we observe that \nf{Patt} achieves again the best precision values for the English corpora. 
On the other hand, choosing the best hypernym worked very well for \nf{DocSub} which obtained the best precision for the Portuguese corpora. 
As filtering out multiple hypernyms might remove also correct relations, the recall values for all corpora are very low. 
Comparing the values achieved by methods containing all relations (Table \ref{tab:top1000}) and the reduced taxonomies, we can see that all recalls and f-measures decreased, as expected. 
The values of precision increased for most corpora of the \nf{Patt} and \nf{DocSub} models. 
Using f-measure as reference, it seems not worth to apply this kind of filtering because the loss in recall is much greater than the gain in precision.

Comparing the values achieved by \nf{HClust} using 1,000 terms and the ones obtained when reducing to one parent, we can see that they are almost the same. 
Only TED Talks corpora obtained a decrease in precision (from $\mathcal{P}$=0.0664 to $\mathcal{P}$=0.0661 in English and from $\mathcal{P}$=0.5656 to $\mathcal{P}$=0.5295 in Portuguese). 
It seems that when applying the hierarchical clustering the taxonomies are reduced as they are when the algorithm to one parent is applied.

Analyzing all results generated for each method in each corpora, the algorithm to choose the best parent seems to work mainly with \nf{Patt} and \nf{DocSub} models. 
Increasing the number of terms form 1,000 to 10,000 only increased the recall in \nf{TF} and \nf{DF} models. 
Clustering semantically related terms also seems a good strategy to reduce the number of relations and increase the precision, when not considering the recall. 
Regarding the corpora, the highest precision obtained by corpora in English was achieved by \nf{Patt} models, while the highest recall by statistical models such as \nf{TF} and \nf{DF}. 
The highest f-measure for corpora in English was achieved by \nf{TF} model in Europarl and \nf{DocSub} model in TED Talks. 
For the corpora in Portuguese, the highest precision was achieved by \nf{DocSub} model, while the highest recall and f-measures were achieved by \nf{TF} models.

In the next section we analyze in detail the metrics of the hierarchies generated by each method.

\subsection{Metrics Analysis}
\label{subsec:metrics_analysis}

Another way to see learned taxonomies is on the basis of their characteristics. 
The characteristics can be translated into a group of metrics that were initially developed for ontology evaluation \cite{Freitas2010,FreitasVieira2010}. 
In this work, we adapted metrics that are used for ontologies and applied them in all learned taxonomies in order to see what types of taxonomies each model generates. 
A description of each metric is presented in Section \ref{subsec:metrics}.

\begin{figure}[tbh]
    \centering
    \includegraphics[width=0.65\textwidth]{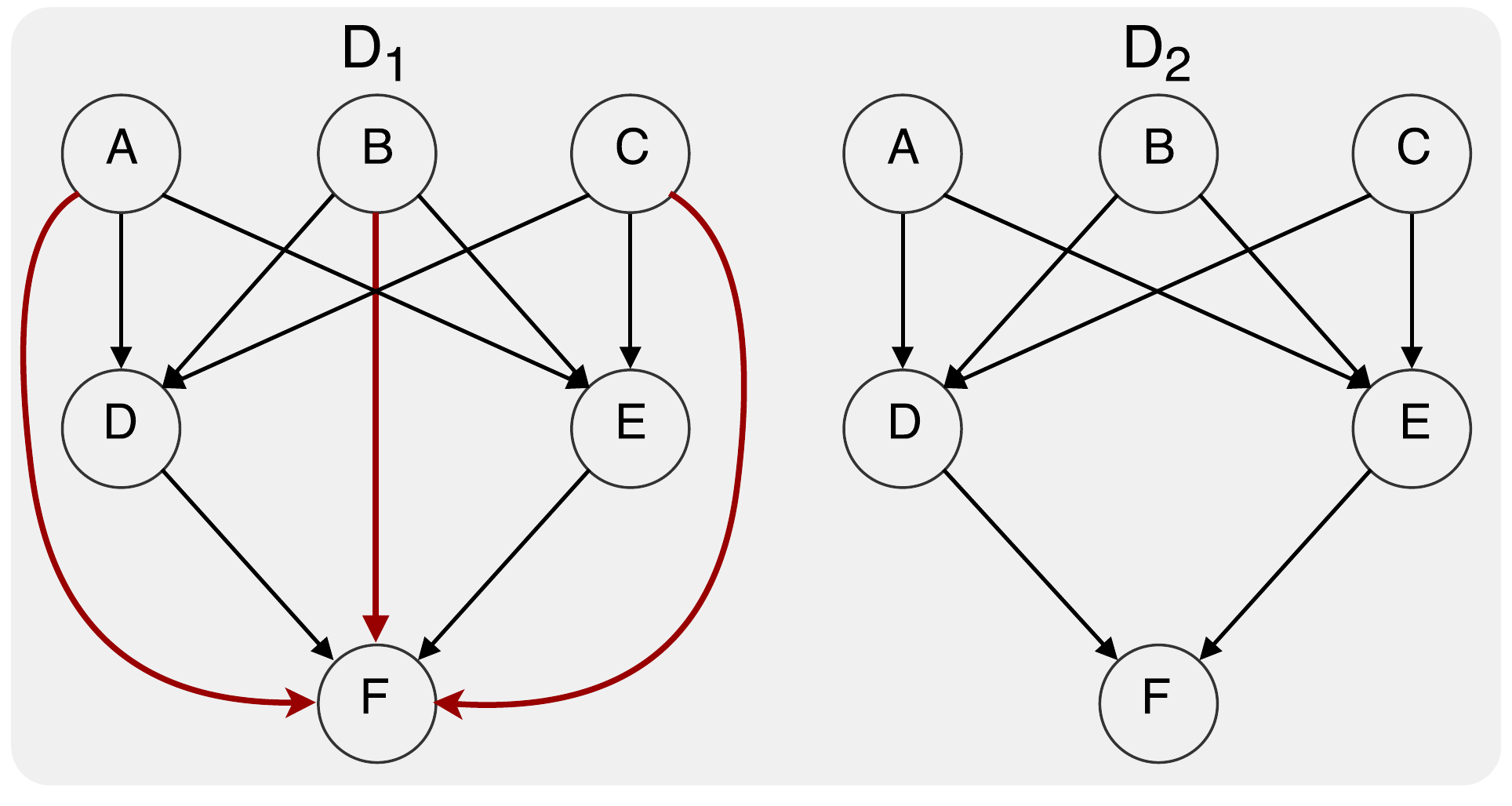}
    \caption{Transitive reduction from digraph \nf{D\txtsub{1}} to \nf{D\txtsub{2}}.}
    \label{fig:graph_transitivity}
\end{figure}

Analyzing values generated by each metric, specifically the maximum depth of models that use the statistics between words to generate the taxonomy (\nf{DSim}, \nf{SLQS}, \nf{TF}, \etc), we observe that the distance between a root term, \idest, a term that does not have a parent term, and a leaf term, \idest, a terms that does not have a child term, was equal to 1, meaning that a direct edge connects the root term and the leaf term, as well as all the other terms. 
For example, consider the digraph D\txtsub{1} in Figure \ref{fig:graph_transitivity}, where all relations between terms are shared using only one edge (\exemp, the distance between the root term  ``A'' and the leaf term ``F'' is equal to 1 because there is a direct edge connecting them). 

To calculate the maximum distance between a root term and a leaf term we decided to remove edges using the transitive reduction of a finite directed graph. 
The transitive reduction generates the minimum equivalent graph, \idest, a graph with the fewest possible edges that has the same reachability relation as the original graph. 
Thus, the minimum graph must contain all paths, but not necessarily all edges, between nodes as the original graph. 
For example, the digraph ``D\txtsub{2}'' presented in Figure \ref{fig:graph_transitivity} shows the transitive reduction of the original digraph ``D\txtsub{1}'', where the edges connecting \nf{A}$\rightarrow$\nf{F}, \nf{B}$\rightarrow$\nf{F} and \nf{C}$\rightarrow$\nf{F} are removed, and all paths between terms are kept. 
For instance, the direct edge connecting terms \nf{A} and \nf{F} are removed, but the path between them pass through \nf{D} or \nf{E} (path \nf{A}$\rightarrow$\nf{D}$\rightarrow$\nf{F} or \nf{A}$\rightarrow$\nf{E}$\rightarrow$\nf{F}), as well as the path between \nf{C} and \nf{F} pass through \nf{D} or \nf{E} (path \nf{C}$\rightarrow$\nf{D}$\rightarrow$\nf{F} or \nf{C}$\rightarrow$\nf{E}$\rightarrow$\nf{F}). 
Besides reducing the number of relations, removing transitivity creates a structure as it is usually presented in semantic relation networks.

Using the direct graph with transitive reduction for each model, we generated the metrics presented in Table \ref{tab:metrics1k_en}. 
The results are generated for models using the top 1,000 terms of each corpus in English generated for the automatic evaluation, \idest, models using nouns instead of noun phrases. 
It is also important to notice that as we generated several taxonomies in \nf{DocSub} with different thresholds, and several taxonomies in \nf{HClust} with different number of clusters, we selected the ones that achieved the highest f-measure value. 
Thus, the taxonomy generated by \nf{DocSub} model using the TED Talks corpus contains the threshold $\lambda$=0.3 and the taxonomy generated using the Europarl corpus contains threshold $\lambda$=0.1. 
For the \nf{HClust} models, the taxonomy contains 1,000 terms and the maximum number of clusters. 
\nf{Patt} model contains relations of all patterns for the limited number of terms.

\begin{table*}[ht]
 \caption{Metrics for taxonomies generated by models using the top 1,000 terms of each corpus in English.}
 \label{tab:metrics1k_en}
 \centering
 \begin{tabular}{c|r|rrrrrrr}
   \hline\noalign{\smallskip}
    \noalign{\smallskip}
     Corpus                     & Metric        & Patt  & DSim  & SLQS  & TF    & DF    & DocSub & HClust \\
    \noalign{\smallskip}\hline\noalign{\smallskip}
    \multirow{10}{*}{Europarl}  & TotalTerms:   & 957   & 1,000 & 1,000 & 1,000 & 1,000 & 836   & 1,000 \\
                                & TotalRoots:   & 44    & 1     & 1     & 1     & 1     & 43    & 1     \\
                                & NumberRels:   & 1,588 & 1,025 & 1,028 & 1,185 & 1,103 & 1,184 & 999   \\
                                & MaxDepth:     & 21    & 921   & 901   & 788   & 835   & 8     & 15    \\
                                & MinDepth:     & 1     & 921   & 901   & 788   & 835   & 1     & 1     \\
                                & AvgDepth:     & 11.82 & 921   & 901   & 788   & 835   & 3.05  & 8.46  \\
                                & DepthCohesion:& 1.78  & 1     & 1     & 1     & 1     & 2.62  & 1.77  \\
                                & MaxWidth:     & 20    & 2     & 3     & 4     & 3     & 88    & 41    \\
                                & MinWidth:     & 1     & 1     & 1     & 1     & 1     & 1     & 1     \\
                                & AvgWidth:     & 1.99  & 1.03  & 1.03  & 1.19  & 1.10  & 4.20  & 2.38  \\
    \noalign{\smallskip}\hline\noalign{\smallskip}
    \multirow{10}{*}{TED Talks} & TotalTerms:   & 476   & 1,000 & 1,000 & 1,000 & 1,000 & 1,000 & 1,000 \\
                                & TotalRoots:   & 164   & 2     & 1     & 1     & 1     & 1     & 1     \\
                                & NumberRels:   & 521   & 1,029 & 1,331 & 3,025 & 3,438 & 3,802 & 1,009 \\
                                & MaxDepth:     & 16    & 915   & 658   & 454   & 395   & 118   & 12    \\
                                & MinDepth:     & 1     & 913   & 658   & 454   & 395   & 110   & 1     \\
                                & AvgDepth:     & 5.82  & 914   & 658   & 454   & 395   & 112.24 & 5.95 \\
                                & DepthCohesion:& 2.75  & 1     & 1     & 1     & 1     & 1.05  & 2.02  \\
                                & MaxWidth:     & 25    & 2     & 77    & 13    & 12    & 66    & 98    \\
                                & MinWidth:     & 1     & 1     & 1     & 1     & 1     & 1     & 1     \\
                                & AvgWidth:     & 1.83  & 1.03  & 1.36  & 3.03  & 3.44  & 6.64  & 2.35  \\
    \noalign{\smallskip}\hline
 \end{tabular}
\end{table*}

As we can observe in Table \ref{tab:metrics1k_en}, limiting the number of terms to 1,000, \nf{Patt} and \nf{DocSub} do not to generate relations for all terms. 
\nf{Patt} model could not generate relations for all terms because terms must to be in a pattern in order to have their taxonomic relation identified. 
As explained before, patterns are very sparse and not all terms would fit in a pattern. 
The size of the corpus also impacts the number of terms that belong to a relation, thus, the larger the corpus, the easier to find terms that match a pattern. 
For \nf{DocSub} model, the limited threshold may have influenced in results. 
As we expect that two terms share at least 30\% of the documents to exist a taxonomic relation between them, then not all terms would have a relation.

An interesting characteristic of \nf{DSim}, \nf{SLQS}, \nf{TF} and \nf{DF} models is the high number of generated relations as explained before, where almost all terms are connected before applying the transitive reduction. 
When the transitive reduction is applied on these models, they transform the taxonomy into a deep taxonomy, where small differences may indicate a taxonomic relation. 
For example, using relations generated by \nf{TF} model using the Europarl corpus, we can understand the \nf{MaxDepth} as having 789 terms with different values of term frequency, while having 211 that share the same value of term frequency with other terms. 
Using the \nf{SLQS} model, we can understand that 902 terms share different values of entropy, while 98 share the same value with other terms. 
For such models, the number of generated relation is very high before applying the transitive reduction (\exemp, \nf{DSim} contained a total of 499,101 relations before applying the transitive reduction, meaning that almost all terms are interconnected by taxonomic relations).

The maximum width (\nf{MaxWidth}) for these models can be understood as the number of terms that share the same characteristic. 
For instance, the \nf{MaxWidth} for \nf{TF} model using the Europarl corpus is equal to 4, meaning that there are at least one term that share its value of frequency with other 4 terms. 
Comparing the values of both corpora we observe that a small corpus tends to have less difference in these characteristics. 
Thus, in a small corpus the maximum depth (\nf{MaxDepth}) tends to be smaller and the maximum width (\nf{MaxWidth}) tends to be greater when comparing with a larger corpus.

As hierarchical clustering (\nf{HClust}) groups semantically similar terms before verifying the taxonomic relation between them, then the number of relations (\nf{NumberRels}) tends to be smaller than the number of relations generated by \nf{DF} model. 
Such clustering also transforms the generated taxonomy into a dense taxonomy since the number of maximum depth and width are closer than the values presented by \nf{DF}.

Table \ref{tab:metrics1k_pt} contains the results for each metric after applying the transitive reduction for relations generated by models using the top 1,000 terms of each corpus in Portuguese. 
As \nf{DocSub} generated a range of taxonomies with different threshold ($\lambda$) values, the results presented use the taxonomies generated with the highest value of f-measure ($\lambda$=0.1 for both taxonomies). 
For \nf{HClust} model the taxonomies contain all 1,000 terms, since both models achieved the highest f-measure with the maximum number of clusters. 
The \nf{Patt} models contain all patterns but the limited number of terms.

\begin{table*}[ht]
 \caption{Metrics for taxonomies generated by models using the top 1,000 terms of each corpus in Portuguese.}
 \label{tab:metrics1k_pt}
 \centering
 \begin{tabular}{c|l|rrrrrrr}
   \hline\noalign{\smallskip}
    \noalign{\smallskip}
     Corpus                     & Metric        & Patt  & DSim  & SLQS  & TF    & DF    & DocSub & HClust \\
    \noalign{\smallskip}\hline\noalign{\smallskip}
    \multirow{10}{*}{Europarl}  & TotalTerms:   & 980   & 1,000 & 1,000 & 1,000 & 1,000 & 996   & 1,000 \\
                                & TotalRoots:   & 79    & 1     & 1     & 1     & 1     & 1     & 1     \\
                                & NumberRels:   & 1,527 & 1,031 & 1,049 & 1,185 & 1,093 & 1,644 & 999   \\
                                & MaxDepth:     & 19    & 902   & 894   & 784   & 849   & 6     & 10    \\
                                & MinDepth:     & 1     & 902   & 894   & 784   & 849   & 1     & 1     \\
                                & AvgDepth:     & 9.43  & 902   & 894   & 784   & 849   & 2.73  & 4.29  \\
                                & DepthCohesion:& 2.02  & 1     & 1     & 1     & 1     & 2.19  & 2.33  \\
                                & MaxWidth:     & 27    & 3     & 3     & 4     & 3     & 201   & 58    \\
                                & MinWidth:     & 1     & 1     & 1     & 1     & 1     & 1     & 1     \\
                                & AvgWidth:     & 1.98  & 1.03  & 1.05  & 1.19  & 1.09  & 6.25  & 2.55  \\
    \noalign{\smallskip}\hline\noalign{\smallskip}
    \multirow{10}{*}{TED Talks} & TotalTerms:   & 296   & 1,000 & 1,000 & 1,000 & 1,000 & 1,000 & 1,000 \\
                                & TotalRoots:   & 101   & 1     & 1     & 1     & 1     & 1     & 1     \\
                                & NumberRels:   & 291   & 1,045 & 1,229 & 3,637 & 4,284 & 2,875 & 999   \\
                                & MaxDepth:     & 10    & 860   & 727   & 388   & 354   & 252   & 17    \\
                                & MinDepth:     & 1     & 860   & 727   & 388   & 354   & 249   & 1     \\
                                & AvgDepth:     & 3.94  & 860   & 727   & 388   & 354   & 250.43 & 6.16 \\
                                & DepthCohesion:& 2.54  & 1     & 1     & 1     & 1     & 1.01  & 2.76  \\
                                & MaxWidth:     & 37    & 3     & 79    & 18    & 13    & 9     & 41    \\
                                & MinWidth:     & 1     & 1     & 1     & 1     & 1     & 1     & 1     \\
                                & AvgWidth:     & 1.79  & 1.05  & 1.23  & 3.64  & 4.29  & 2.94  & 2.37  \\
    \noalign{\smallskip}\hline
 \end{tabular}
\end{table*}

The results for the Portuguese corpora are quite similar to the ones generated by the English corpora, having terms without relations in \nf{Patt} and \nf{DocSub}, and \nf{DSim}, \nf{SLQS}, \nf{TF} and \nf{DF} generating deep taxonomies, affirming the characteristics of each method. 
For Portuguese, the number of relations found by \nf{Patt} model using the TED Talks corpus were smaller than the one found using the English corpus, impacting the maximum depth. 
On the other hand, the number of siblings for a term was greater.

\subsection{Complementarity Analysis}
\label{subsec:complementarity}

\begin{figure}[tbh]
    \centering
    \includegraphics[width=0.85\textwidth]{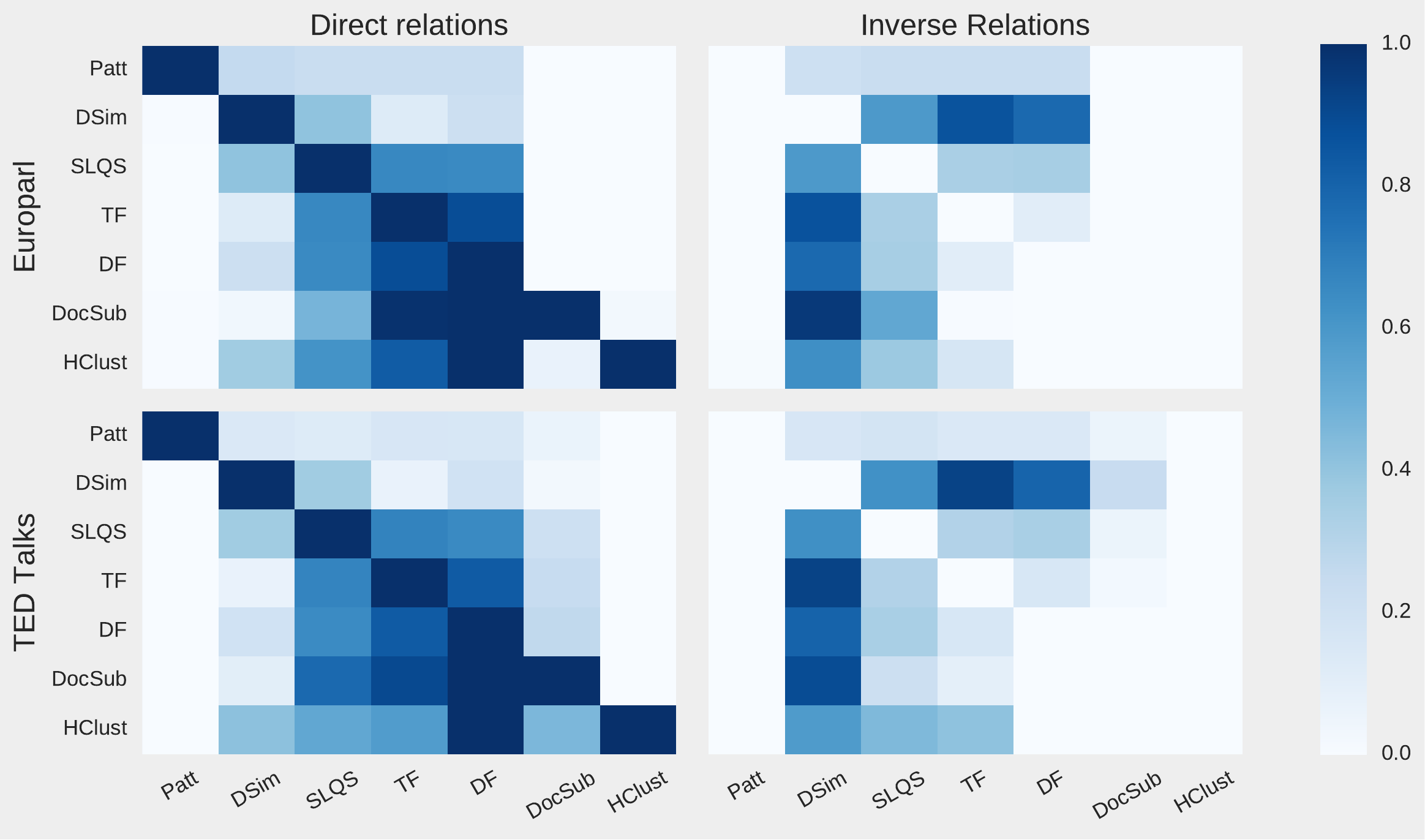}
    \caption{Ratios for relations shared by models in the English corpora.}
    \label{fig:ratio_intersection_en}
\end{figure}

By observing all generated taxonomies, we analyzed how models are complementary, \idest, verifying whether the same relations are generated by more than one model. 
We also observed whether relations are generated in opposite directions (\exemp, \nf{Patt} model generates the relation \nf{A}$\rightarrow$\nf{B} and \nf{DSim} model generates the inverse relation \nf{B}$\rightarrow$\nf{A}). 
To illustrate the effect of mixing models, Figure \ref{fig:ratio_intersection_en} presents a color map with the ratios of direct relations, \idest~relations that are equal in both models, and inverse relations. 
All ratios are generated by models using the top 1,000 word of the English corpora. 
The ratio is computed as the number of taxonomic relations shared by the models in the row and the model in the column divided by the number of relations generated by the model in the row. 
For example, \nf{Patt} model using Europarl corpus in English generated a total of 15,797 taxonomic relations from which 4,014 were shared with \nf{DSim}. 
Thus, the value of the (\nf{Patt}, \nf{DSim}) cell is ($\frac{4,014}{15,797}$)=0.2541. 
It is important to notice that the ratio takes into account the order of the relation, thus, the cell (\nf{DF}, \nf{HClust}) is not equal to the cell (\nf{HClust}, \nf{DF}). 
Each matrix element is indicated with a color, going chromatically from a dark blue for higher values (starting with 1 and found in the diagonal of the direct matrix) to a light blue for lower values (ending with 0 and found in the diagonals of the inverse matrix). 

Such colorful representation let us observe that some models share a high number of relations, while other models share a high number of inverse relations. 
Models that share low number of direct relations and low number of inverse relations tend to be complementary, since relations in one model are not generated in the other model and vice versa. 
Observing the patterns generated by colors in Figure \ref{fig:ratio_intersection_en}, the common point in both corpora is the fact that most models share a high number of relations with \nf{SLQS}, \nf{TF} and \nf{DF} and low number of relations with  \nf{Patt}, \nf{DocSub} and \nf{HClust}. 
Also, a high number of inverse relations are shared with \nf{DSim}. 

Although the number of relations shared with \nf{Patt} are very low (close to zero) for \nf{DSim}, \nf{SLQS}, \nf{TF} and \nf{DF} models, the number of shared relations for \nf{Patt} model is almost 20\% for each model. 
The number of shared relations with \nf{Patt} by \nf{DocSub} and \nf{HClust} is very low in both ways. 
Thus, these models tend to be complementary to the \nf{Patt} model. 
This relation between models is more explicit in the Europarl corpus.

Almost 60\% of the relations in \nf{SLQS} model are shared directly with \nf{TF} and \nf{DF} and almost 30\% of its relations are shared inversely with these models. 
It means that these models generated relations for almost the same terms, and only few relations between terms were not discovered. 
This can be seen by the number of relations found by each model (494,871 for \nf{SLQS}, 498,222 for \nf{TF} and 497,905 for \nf{DF} using the TED Talks corpus). 
As term frequency and context frequency are somehow related, using the top 1,000 terms that share most contexts will create relations for almost all terms. 

\nf{HClust} model have a high number of similar relations with \nf{DF} and \nf{TF} models. 
It totally makes sense for relations shared with \nf{DF} since \nf{HClust} is driven by relations generated by this model, \idest, \nf{HClust} keep only \nf{DF} relations that are semantically related. 
The high number of shared relations with \nf{TF}, mainly in the Europarl corpus, is because \nf{DF} move towards \nf{TF} when the size of the documents decrease. 
As \nf{HClust} is driven by \nf{DF}, it is also affected by the size of the document and consequently it approaches \nf{TF}.

\begin{figure}[t!]
    \centering
    \includegraphics[width=0.85\textwidth]{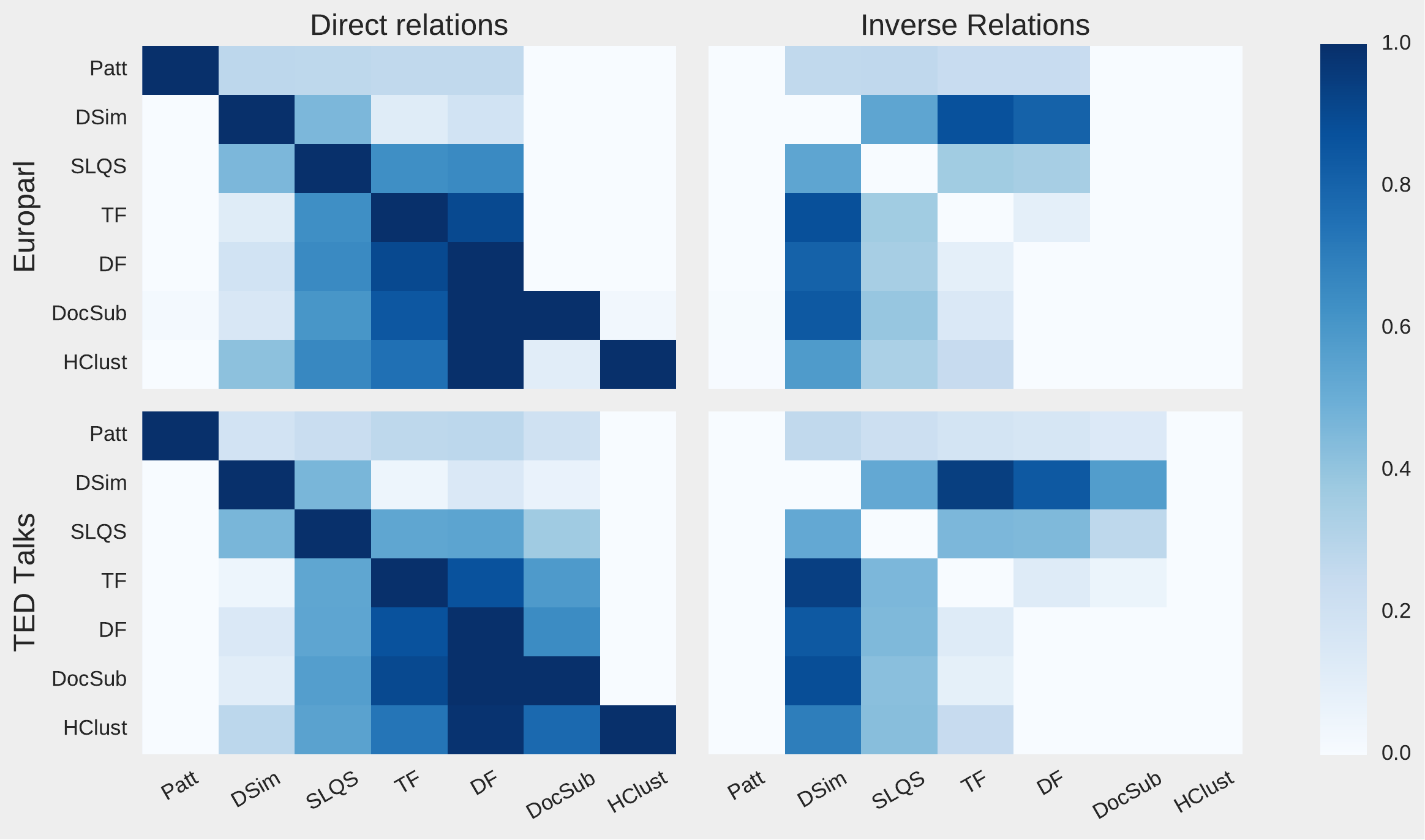}
    \caption{Ratios for relations shared by models in the Portuguese corpora.}
    \label{fig:ratio_intersection_pt}
\end{figure}

Figure \ref{fig:ratio_intersection_pt} shows the color map for models using the Portuguese corpora to generate taxonomic relations. 
As occurred when analyzing metrics, the results generated by these models are quite similar to the results generated using the English corpora. 
The most notable difference between English and Portuguese results rely on the high number of inverse relations generated by \nf{TF} when the source is \nf{DocSub}, when using the Europarl corpus in English. 
Although values in cells are quite similar for the most results, the value in cell (\nf{DocSub}, \nf{TF}) using the Europarl in Portuguese is very low when comparing with the results in English. 

Minor differences may be also observed in \nf{Patt} model which has more shared relations with other models in English than in Portuguese. 
On the other hand, the inverse relation of this model with \nf{DocSub} model using Europarl in Portuguese is higher when compared with the relation using the English version of the corpus.

Observing the relations shared between models, we believe that it would be also interesting to verify the impact in precision when combining models. 
The relative precision ($\mathcal{P}_R$) indicates what is the gain or loss of the model when using only terms that are shared by another model. 
Thus, we calculate the relative precision as the precision of the relations in the intersection of the models divided by the precision of the original model. 
For instance, consider that the precision in the original \nf{Patt} model containing the top 1,000 terms is equal to $\mathcal{P}$=0.53, and the precision taking into account only terms shared with \nf{DocSub} model is equal to $\mathcal{P}$=0.65. 
The relative precision is the ratio between both values $\mathcal{P}_R$=1.2. 
Values above 1 of relative precision indicate that the precision of the relations in the intersection of the models is higher than the precision of the original model. 
Values between 0 and 1 indicate that the intersection between models affect negatively the original model.

\begin{figure}[t!]
    \centering
    \includegraphics[width=0.85\textwidth]{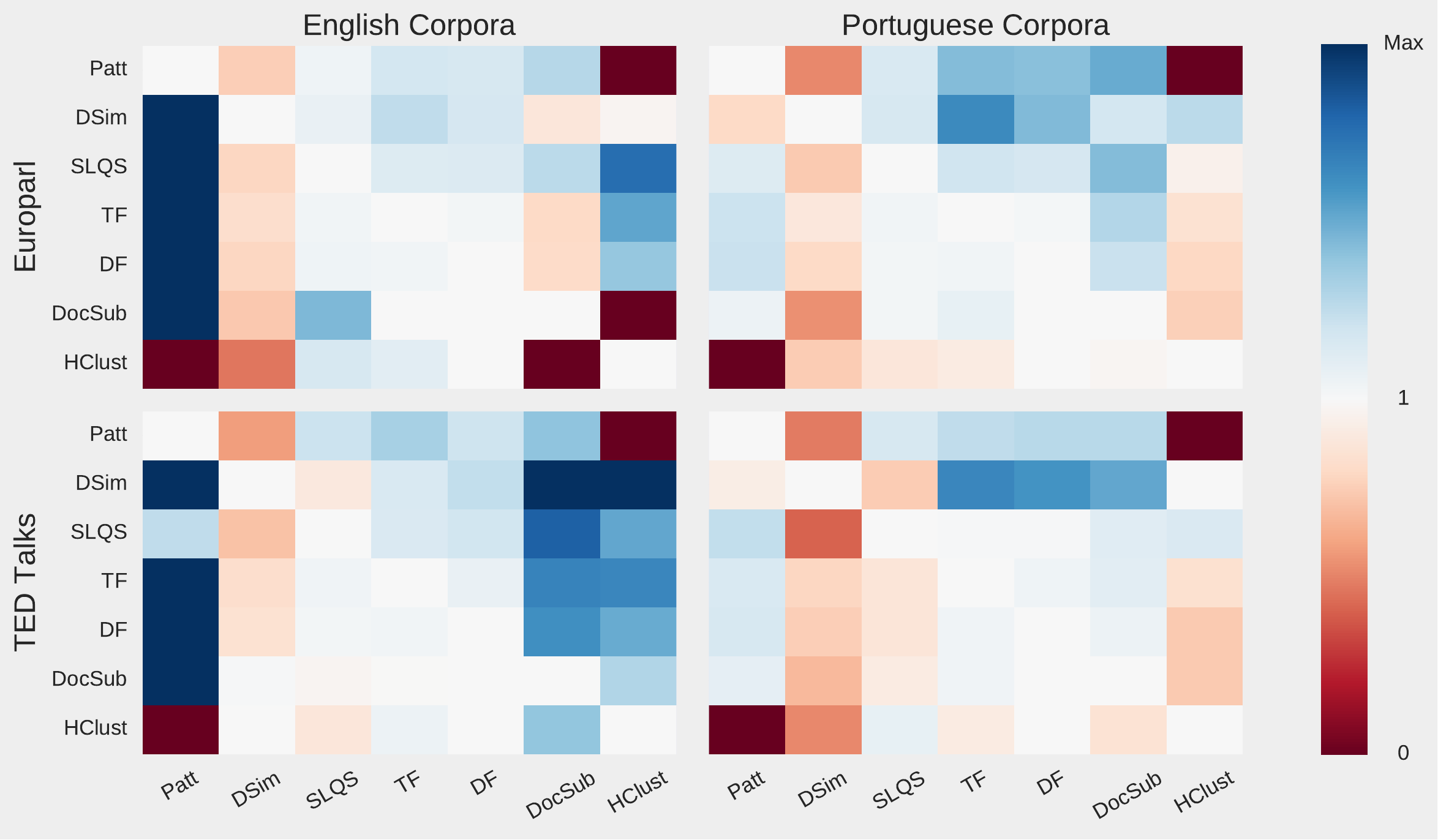}
    \caption{Relative precision using the intersection of the models.}
    \label{fig:shared_precision}
\end{figure}

Figure \ref{fig:shared_precision} illustrates a color map with the relative precision scores achieved by each intersection of the model in each corpus for English and Portuguese. 
All $\mathcal{P}_R$ are generated by models using the top 1,000 word of each corpus. 
It is important to notice that the relations of the original model are the rows of the matrix and the models used in the intersection are the columns. 
Thus, the cell (\nf{DF}, \nf{HClust}) contains the precision of the relations in the intersection divided by the precision of the original model \nf{DF}. 
Each matrix element is indicated with a color, going chromatically from a dark blue for higher relative precision values, passing through white for neutral value of relative precision (values close to 1), to a dark red for lower values of relative precision (values close to zero). 
Thus, if the cell contains a shade of blue the intersection achieves a higher precision when compared to the original and if it contains a shade of red the precision in the intersection is lower than the original. 

As observed in the previous experiment, the number of shared relations between \nf{SLQS}, \nf{TF} and \nf{DF} is very high.
Having this high number of shared relations, the intersection of these models do not significantly change the precision when compared to the their original scores. 
The precision increased for most methods when they are combined with \nf{Patt}. 
The unique exception is the \nf{HClust} model which decreased the precision. 
This low value of precision is due to the fact that there are no relations in the intersection of both methods. 
As the intersection is equal to zero, the value of precision of the intersection is also equal to zero.

Most methods also increased the precision using the intersection with the \nf{DocSub} model. 
On the other hand, most methods decreased when combined with \nf{DSim} model. 
Methods using corpora in English tend to increase the precision when combined with \nf{HClust}, meaning that it seems interesting to generate taxonomic relations only for semantically related terms. 
\nf{DSim} model benefits from the filtering of relations when combining almost all other models. 
This benefit might be due to the low value of precision that \nf{DSim} achieved, and thus, when filtering relations that are incorrect, the model increase the precision. 

Observing generally all results, \nf{Patt} and \nf{DocSub} are models that may serve as filter in order to increase the precision score. 
On the other hand, when a model is combined with \nf{DSim} the values of precision tend to decrease.

\section{Conclusions and Future Work}
\label{sec:conclusion}

This paper presents a detailed comparative evaluation of different methods for automatic taxonomic relation extraction from text corpus, considering variations in genre and language. 
We started our work by rising questions that we would like to answer. 
Our four main questions were: is there a method that outperform all other methods? 
If changing the language, the method perform equally? 
All methods generate similar taxonomies? 
Are results generated by different methods complementary or dissimilar? 
In order to answer these questions, we developed, evaluated and characterized in terms of hierarchy metrics a set of methods that are the state of the art according to the literature in the area.

Ontology learning from texts is an active research field that has produced, since the last decades, a large body of proposals covering the extraction process under different perspectives. 
However, there are many  difficulties in reproducing the results reported in the literature due to the use of different corpora or gold standards. 
While some proposals report improvements on previous approaches, others apply similar strategies on different corpora, making hard to draw a comprehensive comparative between them.
Therefore we decided for developing and evaluating the main models to extract taxonomic relations and evaluate them on the basis of the same corpora. To consider more than one language, we made that analysis over parallel  Portuguese and English texts.

We examined characteristics of the methods for extract taxonomic relations from text corpora by using automatic evaluations. 
The results were measured in terms of precision, recall and f-measure. 
The learned taxonomies were also characterized in terms of hierarchy metrics such as depth, width \etc. 
The intersections of the relations generated by the methods were analyzed in terms of their complementarity or similarity. 

From the automatic evaluation perspective, we confirmed that methods that use the distribution of words in contexts tend to have low values of precision while having higher recall scores, whereas methods using patterns have high values of precision but their recall is very low, due to the scarcity of patterns in texts. 
In order to improve results, it seems a good option to cluster terms before identify the taxonomic relation. 
On the other hand, increasing the number of terms in the model decreases values of precision for most methods with exception of the method that uses patterns and the method that uses document subsumption to identify taxonomic relations. 
It also seems a good option to select the best hypernym for each term using the algorithm proposed by De Knijff \etal~\cite{DeKnijffEtAl2013} since it reduces substantially the taxonomy and improves the precision. 
On the other hand, using this algorithm the recall and f-measure decrease significantly for most methods.

Taxonomies generated by models in the automatic evaluation using the top 1,000 terms with the highest number of contexts are analyzed in terms of hierarchy metrics. 
We observed that models that use the distribution of the words generate taxonomies with much more relations than other methods. 
Almost all terms were related between themselves and the maximum distance between a root term, \idest, a term without hypernym and a leaf terms, \idest, a terms without hyponym was usually equal to one. 
Having a direct connection between the root and the leaf were difficult to measure the maximum distance between them. 
Thus, we applied a transitive reduction in the taxonomy and observed the characteristics of these new taxonomies. 
Hence, the taxonomies generated by distributional methods are much deeper than the other taxonomies. 
On the other hand, they are very narrow, while taxonomies generated by methods such as document subsumption and hierarchical clustering are wider but not so depth.

Finally, methods were analyzed in terms of complementarity, where we could see that most methods generate the same relations learned by models that use the entropy between terms (\nf{SLQS}), models that use the term frequency \nf{TF} and models that use the document frequency (\nf{DF}). 
Also, the method based on patterns \nf{Patt} seems to be complementary to methods that use the document subsumption \nf{DocSub} and hierarchical clustering \nf{HClust}. 
A high number of inverse relations generated by \nf{DSim}  was also noted, \idest, when a model generates the relation between a term and its hypernym and \nf{DSim} model generates the inverse relation. 
When observing the precision of the mixed models, \nf{DSim} seems to benefit when joined to another method, and \nf{Patt} model seems to benefit when joined to the \nf{DocSub} model more than when joined to the \nf{HClust} model.

There are many major directions that this work can be extended. 
First, the set of methods that we have evaluated is not exhaustive and there are a number of other methods that we could not include. 
For example, it would be interesting to include machine learning methods to compare with the developed methods, all using the same test corpora. 
Second, we applied hierarchical clustering on a method that uses the document frequency to indicate a taxonomic relation. 
As semantically group terms with hierarchical clustering seems to improve results when compared with a method without the clustering process, it would be interesting to apply the hierarchical clustering in other methods and observe whether it improves the results. 
Third, a deeper analysis on the complementarity of the methods would also be interesting. 
In this work, we analyze the intersection of pairs of methods, but a deeper analysis using the intersection of more than two methods, or the union of these methods might generate new insights. 
Forth, perform evaluations using manually generated taxonomies as gold standard for both languages using parallel corpora. 
This type of evaluation would help to understand the nuances of the methods in different languages. 
As we used two different gold standards in this work, it was difficult to identify the reasons why a method generates a high precision in one language and not such high precision in the other. 

Finally, consider alternative dimensions for the evaluation. 
We considered only one dimension of evaluation, performing an automatic evaluation against a gold standard. 
Other interesting dimensions of the evaluation could be an in-use evaluation where we measure whether the taxonomy is more relevant for a document collection semantic annotation, for machine translation or any other task. 
Also, performing evaluations using corpora and gold standards in a domain specific task. 
As we used general gold standards, results might not be truly applicable when using a domain specific corpora.

\bibliographystyle{unsrt}
\bibliography{references}

\end{document}